\documentclass[10pt,twocolumn,letterpaper]{article}

\usepackage[pagenumbers]{iccv} 

\usepackage{float}
\usepackage{multirow, tabularx}
\usepackage{graphicx}

\usepackage{inconsolata}

\usepackage{xcolor}

\usepackage{makecell}

\usepackage [english]{babel}
\usepackage [autostyle, english = american]{csquotes}
\MakeOuterQuote{"}

\usepackage{amssymb}
\usepackage{pifont}
\newcommand{\cmark}{\ding{51}}%
\newcommand{\xmark}{\ding{55}}

\usepackage[linesnumbered,ruled,vlined]{algorithm2e}

\usepackage{wrapfig}

\usepackage[accsupp]{axessibility} 

\definecolor{iccvblue}{rgb}{0.21,0.49,0.74}
\usepackage[pagebackref,breaklinks,colorlinks,allcolors=iccvblue]{hyperref}

\def\paperID{12835} 
\def\confName{ICCV}
\def\confYear{2025}

\title{ReasonVQA: A Multi-hop Reasoning Benchmark with Structural Knowledge for Visual Question Answering}

\author{Duong T. Tran$^{1,2}$\\
{\tt\small Duong.Tran@de.bosch.com}
\and
Trung-Kien Tran$^1$ \\
{\tt\small TrungKien.Tran@de.bosch.com}
\and
Manfred Hauswirth$^{2,3}$ \\
{\tt\small manfred.hauswirth@tu-berlin.de}
\and
Danh Le Phuoc$^{2,3}$\\
{\tt\small danh.lephuoc@tu-berlin.de}
\\[1em]
\textsuperscript{1}Bosch Center for AI, Germany
\quad
\textsuperscript{2}Technical University of Berlin
\quad
\textsuperscript{3}Fraunhofer FOKUS
}

\newcommand{\img}[3]{
\begin{figure}[ht]
\centerline{\includegraphics[width=\columnwidth]{#1}}
\caption{#2}
\label{fig:#3}
\vspace{-0.25cm}
\end{figure}
}

\newcommand{\imgbig}[3]{
\begin{figure*}
\includegraphics[width=\textwidth]{#1}
\caption{#2}
\label{fig:#3}
\vspace{-0.25cm}
\end{figure*}
}

\newcommand{\imgfix}[3]{
\begin{figure}[H]
\centerline{\includegraphics[width=\columnwidth]{#1}}
\caption{#2}
\label{fig:#3}
\end{figure}
}

\newcommand{\imgbigscale}[4]{
\begin{figure*}
\centerline{\includegraphics[width=#4\textwidth]{#1}}
\caption{#2}
\label{fig:#3}
\end{figure*}
}

\newcommand{\hl}[1]{#1}

\newcommand{\revqa}{ReasonVQA\xspace}

\begin{document}
\maketitle

\begin{abstract}
In this paper, we propose a new dataset, \revqa, for the Visual Question Answering (VQA) task. Our dataset is automatically integrated with structured encyclopedic knowledge and constructed using a low-cost framework, which is capable of generating complex, multi-hop questions. We evaluated state-of-the-art VQA models on \revqa, and the empirical results demonstrate that \revqa poses significant challenges to these models, highlighting its potential for benchmarking and advancing the field of VQA. Additionally, our dataset can be easily scaled with respect to input images; the current version surpasses the largest existing datasets requiring external knowledge by more than an order of magnitude.
\footnote{ReasonVQA homepage: \href{https://duong-tr.github.io/ReasonVQA/}{https://duong-tr.github.io/ReasonVQA}}
\end{abstract}
\section{Introduction}
\label{sec:intro}
\img{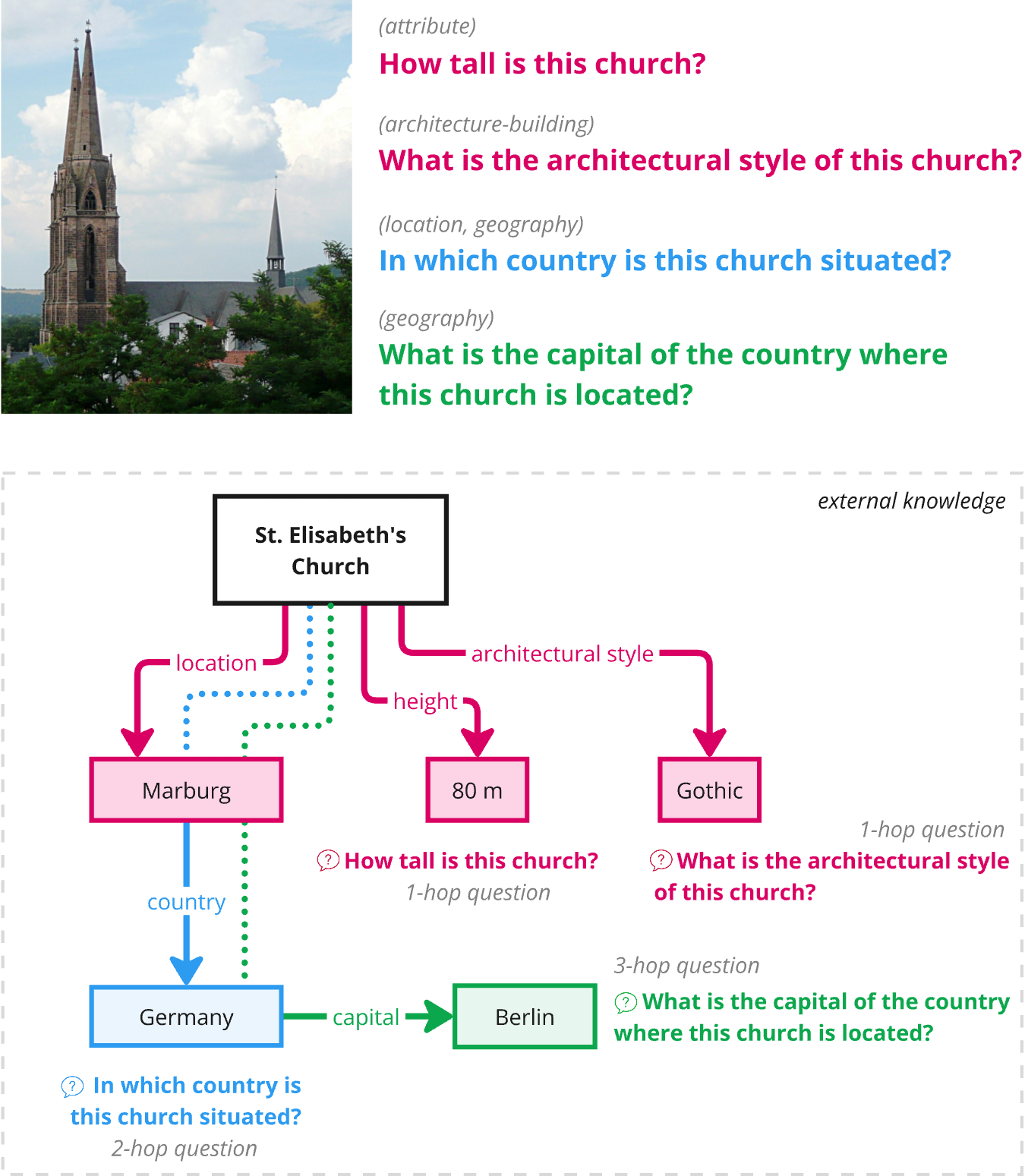}{Sample image and questions from \revqa. Using an existing image, a question is formulated by reasoning through one or multiple hops over the knowledge graph. The generated questions span diverse domains.}{img-intro}

In recent years, significant advancements have been made in the field of Visual Question Answering (VQA) on standard VQA datasets~\cite{DBLP:conf/cvpr/00010BT0GZ18, DBLP:conf/cvpr/HuSDR20, DBLP:journals/pr/YuZWZHT20, DBLP:conf/semweb/0007CGPYC21}. 
Initially, these datasets focused mainly on simple questions related to object identification and attributes, such as name, shape, color, and position. Towards the goal of general-purpose artificial intelligence, VQA models are expected to answer questions that require a deeper understanding of the world, fine-grained visual recognition, and multi-step reasoning. Recently, several additional VQA datasets~\cite{DBLP:conf/cvpr/JohnsonHMFZG17, DBLP:journals/pami/WangWSDH18, DBLP:conf/cvpr/HudsonM19, DBLP:conf/cvpr/MarinoRFM19, DBLP:conf/aaai/ShahMYT19, DBLP:conf/nips/GaoWBP23} have been introduced to challenge VQA systems to handle more complex questions. However, there are limitations associated with these datasets. Some datasets are entirely synthetic, while others rely heavily on manual human effort.
In this paper, we propose a new dataset called \revqa, which was developed from our low-cost and scalable framework. \revqa focuses on integrating external (world) knowledge associated with objects in the images and multi-hop reasoning. For example, in Figure \ref{fig:img-intro}, the question \emph{"How tall is this church?"} requires not only the identification of the church but also additional specific facts, i.e. its height. And the question \emph{"What is the capital of the country where this church is located?"} additionally requires multi-hop knowledge. Such information is often dispersed across multiple paragraphs in the training text, presenting considerable challenges for VQA models.

Our main contributions are summarized as follows: (1) We introduce a new high quality VQA benchmark, called \revqa, which consists of multi-hop questions that require external knowledge to answer;
(2) We propose a scalable, low-cost framework for the construction of \revqa without involving too much manual effort. The largest version of \revqa comprises 4.2 million questions, making it one to two orders of magnitude larger than similar existing datasets. The quality of our dataset is verifiable via a user study. Moreover, the framework is designed to be scalable with respect to the number of input images; and, (3)  We evaluated various state-of-the-art VQA models, including those based on foundation models, on \revqa. Our experiments demonstrate that \revqa presents substantial challenges for both current VQA models and multimodal methods, highlighting its potential for benchmarking and advancing the field of VQA.

\section{Related Work}
Standard VQA datasets~\cite{DBLP:conf/nips/MalinowskiF14, DBLP:conf/cvpr/HuSDR20, DBLP:journals/pr/YuZWZHT20, DBLP:conf/semweb/0007CGPYC21} usually focus on the explicit content in the input images with different complex levels of the questions. In this section, we focus mainly on related works that incorporate external knowledge.

Over the last few years, more datasets with external knowledge have been proposed. Noticeably, KVQA~\cite{DBLP:conf/aaai/ShahMYT19} contains questions that require multi-entity, multi-relation, and multi-hop reasoning over a large knowledge graph. The persons' information and images were collected from Wikidata~\cite{DBLP:journals/cacm/VrandecicK14}, however, they were filtered and identified by human annotators. Marino et al.~\cite{DBLP:conf/cvpr/MarinoRFM19} provided OK-VQA, which includes only questions that require external resources for answering them.
OK-VQA used random images from the COCO dataset~\cite{DBLP:conf/eccv/LinMBHPRDZ14} as the input for two rounds of crowdsourcing and the final dataset contains more than 14K images and questions. While OK-VQA is well-balanced and covers a variety of knowledge categories, its construction process is time-intensive and demands considerable human effort. Moreover, there is no control over the reasoning complexity, i.e. the number of hops in questions.

To tackle the issue of relying on human annotations, Gao et al.~\cite{DBLP:journals/pami/GaoWSC23} proposed a commonsense-based VQA system for templated-based question generation, called CRIC. Starting from dynamic templates, questions are generated automatically using functional programming with the fusion of the scene graph and the commonsense facts from the external knowledge graph. The dataset utilizes the images of Visual Genome~\cite{DBLP:journals/ijcv/KrishnaZGJHKCKL17} with their corresponding Scene Graph annotations. In contrast to our dataset containing specific and detailed knowledge, CRIC only extracted the commonsense facts  mostly from ConceptNet~\cite{DBLP:conf/aaai/SpeerCH17}. Recently, LORA~\cite{DBLP:conf/nips/GaoWBP23} was introduced as a dataset that requires formal and complex Description Logic reasoning. It includes 200,000 automatically generated questions. However, the dataset is confined to food and kitchen scenes, with synthetic images. As a result, it is challenging to extend LORA to new domains of knowledge or to leverage various existing image sources.

\hl{Recent efforts to incorporate external knowledge into datasets, like INFOSEEK~\cite{DBLP:conf/emnlp/ChenHLSCRC23} and Encyclopedic-VQA~\cite{DBLP:conf/iccv/MensinkUCGCZSAF23}, consistently reveal the high cost and limited scalability of human annotation. While automation enables scale, these projects show significant human oversight remains crucial for data reliability. For instance, INFOSEEK relied on semi-automation due to cost, and Encyclopedic-VQA's automated output still required extensive human validation. This underscores the enduring need for labor-intensive quality control, especially for complex or nuanced information.}

\section{The ReasonVQA Dataset}
The \revqa dataset is created to assess the capabilities of VQA models to answer multi-hop questions and to utilize external knowledge. It contains a large number of questions categorized into three levels of complexity: 1-hop, 2-hop, and 3-hop. To answer these questions, models must possess a robust ability to reason across multiple external facts. The overview of our framework, illustrated in Figure~\ref{fig:img-overview}, to create \revqa consists of three steps: (1) External Knowledge Integration, (2) Question Generation, (3) Dataset Construction. In the External Knowledge Integration step (Sect.\ref{sec:kbintegration}) we "link" annotated objects in the images to corresponding entities in external knowledge graphs. In the Question Generation step, multiple-hop questions are generated semi-automatically by traveling through the knowledge graph (Sect.\ref{sec:qg}). In the Dataset Construction step (Sect.\ref{sec:dc}), the question distribution is balanced to reduce bias, and all the images and questions are split into train set and test set with a similar answer distribution. The detailed workflow are described as algorithms in the supplementary material.

\imgbig{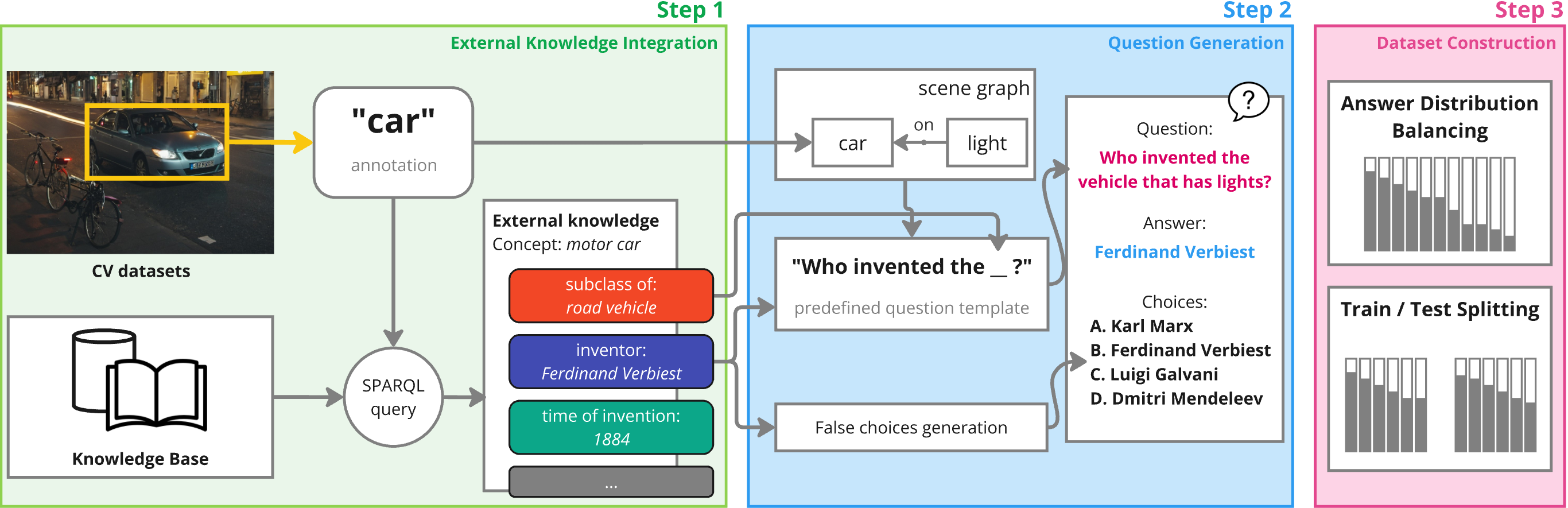}{Overview of the question generation process. The multiple choice question and answer are generated given the input image and the knowledge base. Our approach first utilizes the annotation from the dataset to link the object from the input image to a concept in knowledge base, then assembles the question from a predefined template, and finally generate the multiple choices.}{img-overview}

\subsection{External Knowledge Integration}
\label{sec:kbintegration}

Although our framework is adaptable to any knowledge base and annotated vision dataset, we chose Wikidata~\cite{DBLP:journals/cacm/VrandecicK14}, one of the most complete structure knowledge bases, as the external knowledge source.
We utilize SPARQL to seamlessly integrate image sources, such as Visual Genome (VG)~\cite{DBLP:journals/ijcv/KrishnaZGJHKCKL17}, which contains over 108K images along with various descriptions of objects and their relationships within images, e.g, region descriptions, objects, attributes, relationships, region graphs, scene graphs, and question-answer pairs. VG has been used as a main resource for construction of many other VQA datasets like GQA~\cite{DBLP:conf/cvpr/HudsonM19} and CRIC~\cite{DBLP:journals/pami/GaoWSC23}.
In this paper, we leverage such existing images, questions, object descriptions and scene graphs from VG to build a more robust question generation process. By representing the relationship between an object and other elements, scene graphs allow for a more comprehensive interpretation of the image, avoiding the loss of contextual information from the visual component.
The annotations of objects in VG are canonicalized to WordNet~\cite{DBLP:journals/cacm/Miller95} synset names, which can be used to retrieve the corresponding concepts from Wikidata by using Natural Language Toolkit (NLTK)~\cite{DBLP:books/daglib/0022921} and SPARQL queries. 

Wikidata has rich information about well-known entities such as landmarks and public figures. Hence, we choose Google Landmark Dataset v2 (GLDv2)~\cite{DBLP:conf/cvpr/WeyandACS20} as another image source to expand the knowledge about spatial objects in our candidate images. GLDv2 is a benchmark for large-scale, fine-grained instance recognition and image retrieval in the domain of human-made and natural landmarks. The dataset consists of over 5M images and 200k distinct instance labels to mix with VG. Since the images are annotated with Wikimedia URLs enabling us to retrieve  Wikidata's knowledge, e.g linked concepts and relationships to such URLs. From such URLs, we can extract the landmark names, and the corresponding knowledge in Wikidata.

\subsection{Question Generation}
\label{sec:qg}

\img{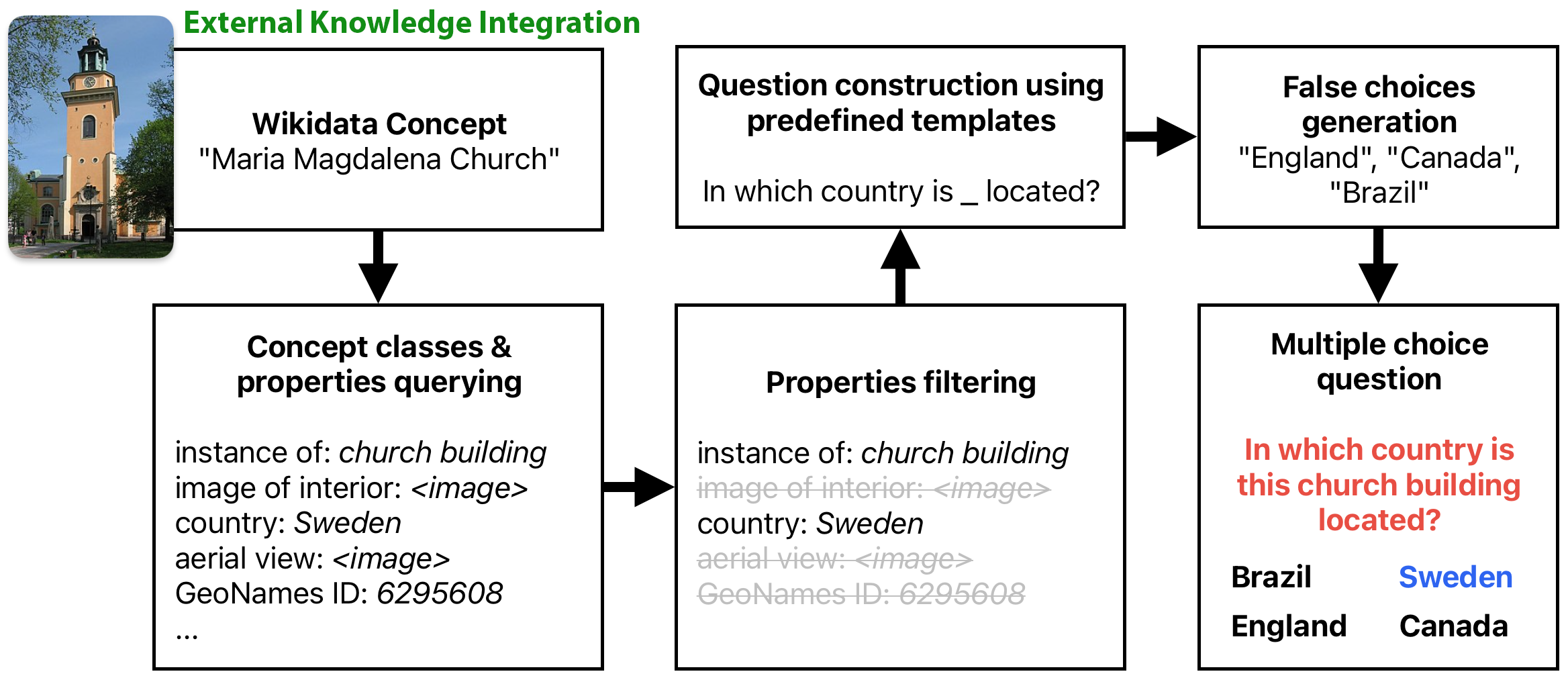}{The process of question generation step-by-step from a Wikidata concept to the multiple choice question. The class hierarchy and all properties of the concept are queried and filtered down to a subset. Then each of the remaining properties is used to generate a multiple choice question.}{img-qg}

\subsubsection{Template-based Question Generation}
\label{sec-tem-qg}

Figure \ref{fig:img-qg} shows the overview of our question generation process. 
\hl{We generate questions through a template-based approach, utilizing predefined templates with fillable placeholders. This method supports the creation of low-cost, scalable systems by enabling a single template to yield many questions. Although the templates are manually crafted, they are designed to be generic enough to provide extensive coverage. To boost linguistic diversity, we also develop linguistic variations for each template. A key advantage of our proposed method is its ability to combine multiple templates into a single question, further enhancing system scalability.}
Furthermore, as we construct potential templates based on knowledge gleaned from the knowledge base (KB), the diversity of the templates aligns with the breadth of knowledge available in the KB.

\textbf{One-hop Question Generation.} Let us assume we are constructing questions around an object in an image, referred to as the \emph{main object}. \hl{This main object corresponds to a Wikidata entity, from which we extract factual knowledge to generate questions.}
After obtaining external facts from Wikidata for the main object as described in Section~\ref{sec:kbintegration}, we use a predefined template for each \emph{property} (or \emph{relation}, interchangeably) occuring in these facts.
Each template contains a placeholder, which will be filled later by the \emph{class name} of the main object, resulting in 1-hop question. For Wikidata, the class name of an object is simply the value of either the "instance of" or "subclass of" property. The Figure \ref{fig:img-hop} shows an example of 1-hop template question. In this case, the template is "Who designed \_\_?" and the class name "skyscraper" is used to fill in the placeholder, producing the complete question "Who designed this skyscraper?". Note that during the process of retrieving facts about the main object from Wikidata, we also obtain the answer to the question. For example, the answer for the mentioned question is "César Pelli".

\textbf{Multi-hop Question Generation.} Multi-hop questions about a main object are constructed by filling the placeholders in 1-hop question templates, which we will now refer to as the \emph{main template}, with a sub-clause. This sub-clause is formed using a \emph{sub-clause template}, which is also manually designed for each relation. For example, the sub-clause template for the relation "architect" is "the architect of \_\_". The placeholder in a sub-clause template can be filled by either the class name of the main object or another sub-clause. This is the cornerstone of our multi-hop question construction mechanism. We additionally leverage the scene graph annotations in VG to construct a sub-clause, e.g. "\_\_ parked next to the sidewalk". This enables us to better integrate semantic visual information into the questions.
Figure~\ref{fig:img-hop} show examples of the templates and the corresponding questions for 2-hop and 3-hop questions.

Additionally, we categorize each question into one or more domains based on the property used to generate the question. For instance, a question about the birthday of a famous person, derived from the property "date of birth," can be classified under both the "person" and "history-events" domains. This enables the system to receive input as one or multiple specified domains and generate questions accordingly. This mechanism is particularly useful for narrowing down the dataset to a specific domain, facilitating more focused and refined results. For this, we predefined 20 domains that encompass all aspects of human knowledge. A detailed list of these domains can be found in Section 2 of the supplementary material.

\imgbig{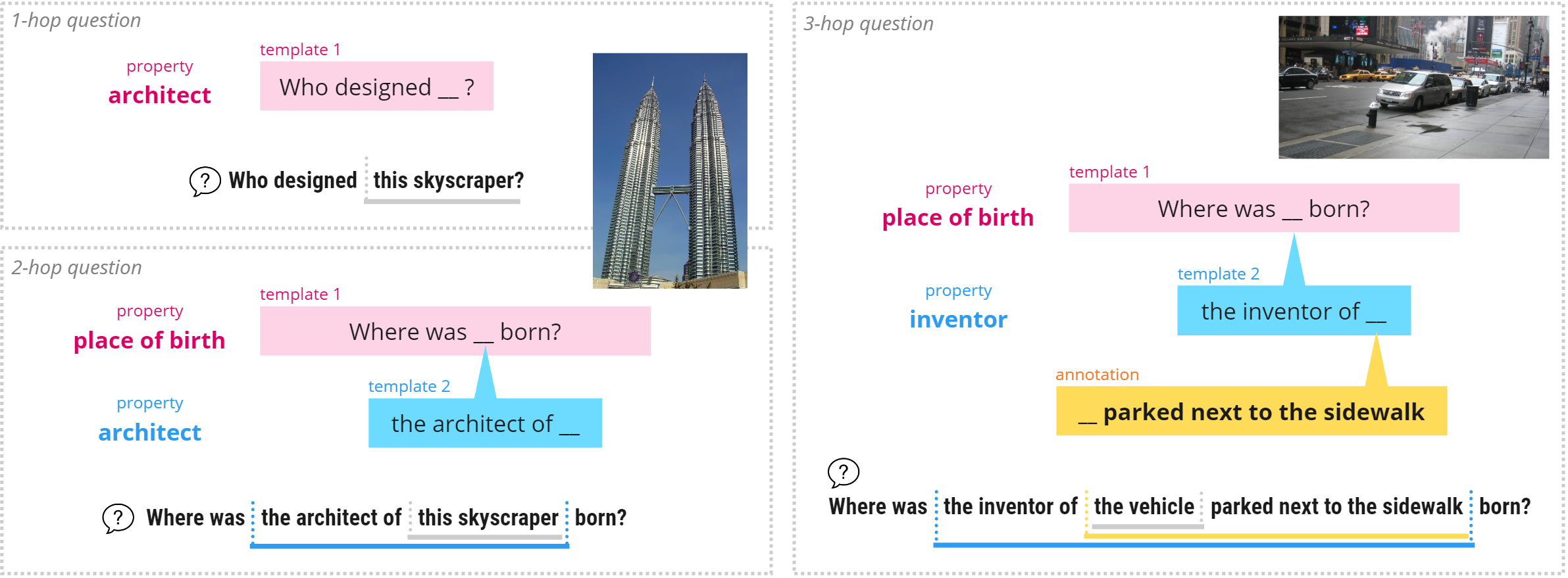}{Example of constructing questions using template-based approach. A 1-hop question is constructed by filling the class name of the main object into a template. Meanwhile, 2-hop and 3-hop questions are constructed in a similar manner by nesting two or more templates.}{img-hop}

\subsubsection{False Choice Generation}
To generate challenging multiple choice questions, in addition to the correct answer, we need to create non-trivial false choices. First, we assign a question into one of four categories based on its correct answer: \emph{fixed}, \emph{date}, \emph{number}, and \emph{literal}. The \emph{fixed} category contains questions that have an enclosed set of choices, such as questions about gender, continent, country, and language. The \emph{date}, \emph{number} and \emph{literal} categories contain questions about date time, numeric values, and literal values, respectively. 
 For the questions in the \emph{fixed} categories, the false choices are randomly selected from the corresponding set of choices. For the \emph{date} category, the false choices are generated by randomly selecting $N$ date values within $[-10, +10]$ years. For the \emph{number} category, the false choices are generated by randomly selecting $N$ numbers within a specific range $[m, n]$, where $m=\frac{i}{2},$ $ n=max(1.5i,$ $ m+2N)$ with $i$ being the correct answer. For the \emph{literal} category, the false choice is generated by retrieving $N$ values that come from the same property as the correct answer. In practice, we set $N=3$, meaning that there is up to three false choices for each correct answer.

\subsection{Dataset Construction}
\label{sec:dc}
\subsubsection{Answer Distribution Balancing}
\label{sec-balancing}

One of the prevalent problems of existing VQA datasets is the strong bias toward specific answers that allows ML models to assume the output without completely understanding the input. Therefore, we look at how to smooth out the answer distribution to alleviate this issue. Inspired by GQA~\cite{DBLP:conf/cvpr/HudsonM19}, we compute the distribution of all the true answers for each group, where a group is defined by its corresponding property. 
We sort the answers by the number of occurrences and select only the top ten most frequent answers in each group for the next step. Then, we create a smoother answer distribution using the following process: We start by going through the answers in a specific group in decreasing frequency order and repeatedly remove ones, together with their questions, to make the \emph{head} size comparable to the \emph{tail} size. The \emph{head} and \emph{tail} are respectively the group of questions having the most and the least number of answers. Essentially, answers with higher frequency will be discarded more often. Thus, this process helps the benchmark become less biased, more balanced and also more challenging to VQA models. While repeating this operation, we maintain the frequency order by keeping minimum and maximum ratios between each pair of consecutive answers. We also prioritize removing answers from images with the most number of questions. Doing that helps us to equalize the number of questions between images and avoid discarding images as much as possible. The balancing process is applied in an iterative manner, with the answer distribution becoming more and more uniform after each iteration. The visualization of our balancing process is provided in Section 1.3 of the supplementary material.

\subsubsection{Dataset Splitting}
\label{sec-splitting}

Since it is necessary to maintain the answer distribution between the train set and test set, the dataset is split through the following procedure: We distribute images into categories and then divide each category randomly by the same train/test ratio. For each image, we first compute the frequency of every answer and select answers that appear at least 1\% in the answer pool, which is the set of all the answers from the dataset. Next, we assign each image to a category, where the category name is constructed by concatenating its top two most frequent answers. We also combine all the categories containing only a single image. Finally, images in each category are split randomly into 70\% train and 30\% test, together with their question-answer pairs. The resulting sets are guaranteed not to have overlapping images, and all the questions about a given image appear in the same split. Please refer to Section 1.4 in the supplementary material for visualization depicting the similarity in answer distribution between the two sets.

\section{Dataset Analysis and Experiments}
\subsection{Dataset Statistics and Quality Evaluation}
The latest version of ReasonVQA consists of more than 598K images and about 4.2M questions. 
\hl{The scalability of our dataset depends on two key factors: (1) numerous objects share common attributes; and (2) multi-hop questions are constructed by nesting multiple templates. For instance, since many facts about building share the "height" attribute, we only need to define a single template for this property during question generation.}
To assess state-of-the-art approaches, we derived a subset version of our dataset, ReasonVQA-U. The main objective is to create a manageable subset of our dataset to assess and compare these models efficiently. The answer distribution balancing process was applied on this subset to obtain the balanced version ReasonVQA-B. 

\begin{table}[tb]
  \centering
  \begin{tabular}{lccc}
    \toprule
    \textbf{Dataset} & \textbf{\#Images} & \textbf{\#Questions} & \textbf{KB} \\
    
    \midrule
    DAQUAR~\cite{DBLP:conf/nips/MalinowskiF14} & 1.4K & 12K & \xmark \\
    COCO-QA~\cite{DBLP:conf/nips/RenKZ15} & 69K & 117K & \xmark \\
    VQA v2~\cite{DBLP:conf/cvpr/GoyalKSBP17} & 204K & 1.1M & \xmark \\
    CLEVR~\cite{DBLP:conf/cvpr/JohnsonHMFZG17} & 100K & 999K & \xmark \\
    GQA~\cite{DBLP:conf/cvpr/HudsonM19} & 113K & 22M & \xmark \\
    PQA~\cite{DBLP:conf/cvpr/QiZSS21} & 32.9K & 157K & \xmark \\
    \midrule
    OK-VQA~\cite{DBLP:conf/cvpr/MarinoRFM19} & 14K & 14K & \cmark \\
    KVQA~\cite{DBLP:conf/aaai/ShahMYT19} & 24K & 183K & \cmark \\
    CRIC~\cite{DBLP:journals/pami/GaoWSC23} & 96K & 494K & \cmark \\
    LORA~\cite{DBLP:conf/nips/GaoWBP23} & 100K & 200K & \cmark \\
    InfoSeek~\cite{DBLP:conf/emnlp/ChenHLSCRC23} & 8.9K & 1.35M & \cmark \\
    Encyclopedic VQA~\cite{DBLP:conf/iccv/MensinkUCGCZSAF23} & 514K & 1M & \cmark \\
    
    \midrule    
    
    ReasonVQA-U & 13.3K & 78K & \cmark \\
    ReasonVQA-B & 8.7K & 51.9K & \cmark \\
    ReasonVQA Full & 598.5K & 4.2M & \cmark \\
  \bottomrule
  \end{tabular}
  \caption{Comparison between ReasonVQA and previous VQA datasets. The top section contains VQA datasets without knowledge component. The middle and bottom sections contains knowledge-based VQA datasets.
  }
  \label{table:ds-compare}
  \vspace{-0.5cm}
\end{table}

In Table \ref{table:ds-compare}, we compare our dataset with various VQA datasets. 
In terms of the number of images, ours has the highest quantity. Moreover, we have a significantly larger number of questions compared to all the others, except GQA. 
To assess the quality of our dataset, we conducted a user study over 1000 randomly selected question and image pairs. The study involves 20 participants, who use English as the main working and/or studying language, to evaluate \hl{the correctness of the answers and naturalness} of 50 randomly selected questions each.  They were asked to rate the naturalness on a four-level scale: (1) very unnatural; (2) unnatural; (3) natural; and (4) very natural. Additionally, participants also marked questions with grammatical errors. \hl{The user study showed that $96\%$ of them are correct. The primary sources of error comes from numerical values, e.g. the area of cities, and ambiguities in the height or length of certain constructions.} We also observed that 2.20\% of the questions were rated as "very unnatural," 13.9\% as "unnatural," 58.1\% as "natural," and 25.8\% as "very natural," with only 2.5\% containing grammatical errors. This user study indicates that more than 83\% of the questions are natural to humans, demonstrating that our dataset already has a good level of naturalness but still has room for improvement. Additional dataset statistics and results of the user study can be found in the Section 2 of the supplementary material.

\hl{
We categorize each question into one or more domains based on the property used to generate the question. For instance, a question about the birthday of a famous person, derived from the property “date of birth,” can be classified under both the “person” and “history-events” domains.
This enables the system to receive input as one or multiple specified domains and generate questions accordingly.
We predefined 20 domains covering a wide range of human knowledge. During question generation, we select only properties associated with the queried domains, filtering out others to optimize efficiency.
Figure~\ref{fig:img-breakdown} visualizes the domain distribution. The complete list of domains is available in Section 2 of the supplementary material.
}

\img{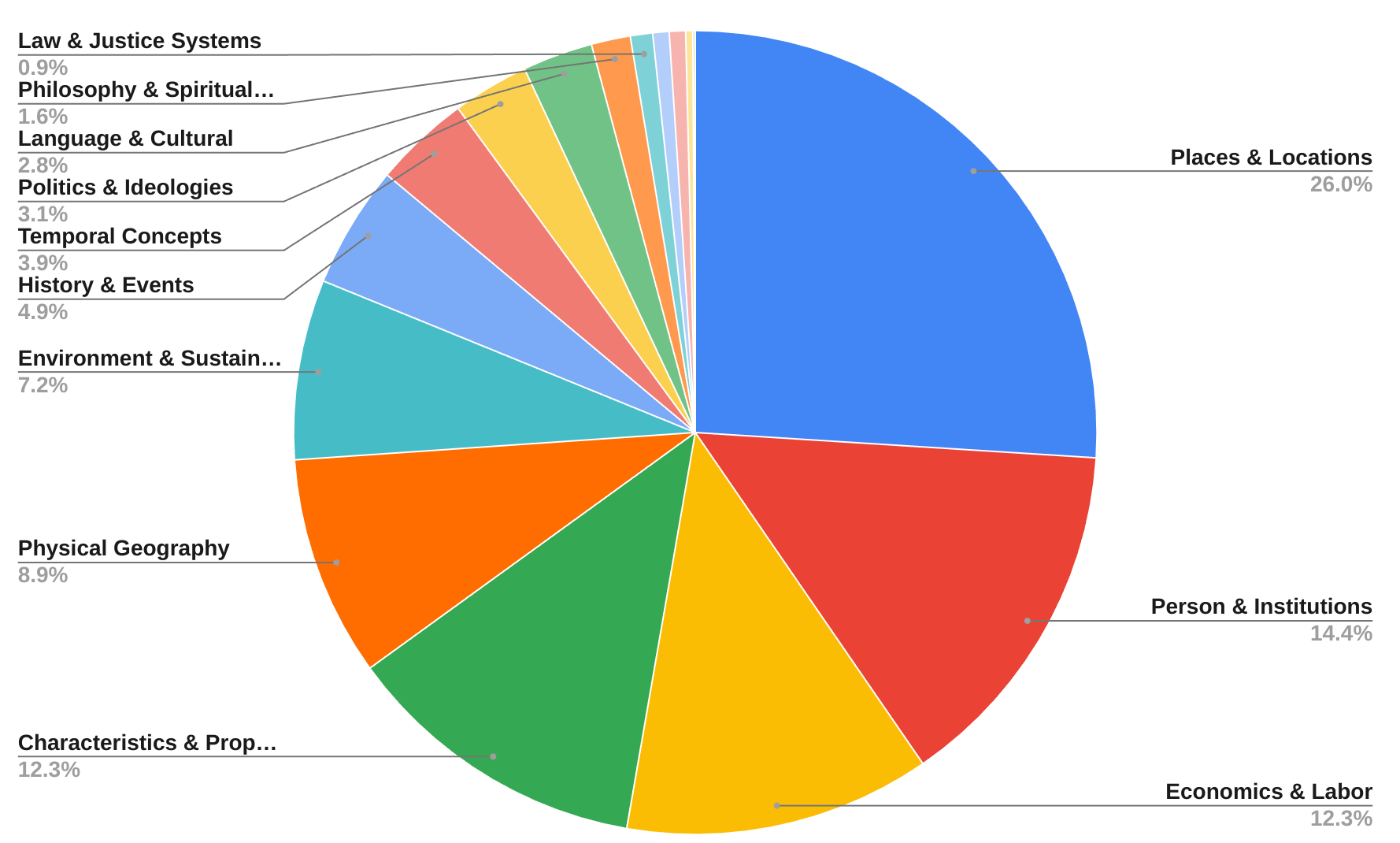}{Distribution of questions across domains. Each question is categorized into specific domains based on the property from which it was generated.}{img-breakdown}

\subsection{Baseline Experiments and Analysis}
\label{sec:experiments}
To rigorously evaluate the challenges posed by our dataset in knowledge-based visual question answering tasks, we perform comprehensive assessments using both the unbalanced and balanced versions of our dataset across a wide range of state-of-the-art vision-language models: BLIP-2~\cite{DBLP:conf/icml/0008LSH23}, InstructBLIP~\cite{DBLP:conf/nips/Dai0LTZW0FH23}, mPLUG-Owl2~\cite{DBLP:journals/corr/abs-2311-04257}, Idefics2~\cite{DBLP:conf/nips/LaurenconTCS24}, Mantis-SigLIP~\cite{DBLP:journals/tmlr/JiangHZWKLC24}, Mantis-Idefics2~\cite{DBLP:journals/tmlr/JiangHZWKLC24}, mPLUG-Owl3~\cite{DBLP:journals/corr/abs-2408-04840}, GPT-4o~\cite{DBLP:journals/corr/abs-2410-21276}, LLaVA-OV~\cite{DBLP:journals/tmlr/0080ZGZ00ZZL0L25}, Qwen2.5-VL~\cite{DBLP:journals/corr/abs-2502-13923}, PaliGemma-2~\cite{DBLP:journals/corr/abs-2412-03555}, PaliGemma-2-Mix~\cite{PaliGemma2Mix2025}, SmolVLM-Instruct~\cite{marafioti2025smolvlm}.

\textbf{Zero-shot Evaluation and Dataset Comparison.} We assessed the zero-shot performance of all models across two scenarios: open-ended and multiple choice. For each model, we constructed prompts directly using the questions from ReasonVQA, with the associated images serving as the input. In open-ended scenario, we collect and evaluate the text answers generated by each LLM. For the multiple-choice setup, we carefully design prompts to instruct each LLM to respond with only a single letter that corresponds to its predicted choice. 

\textbf{Fine-tuning and Performance Gains.} To assess the effectiveness of fine-tuning, we compared the performance of three vision-language models of varying scales before and after training on our dataset. Models were fine-tuned using direct image, question, and answer inputs. This experiment evaluates how well models can adapt to the dataset and improve their understanding of knowledge-based visual question answering. By analyzing performance gains, we demonstrate the potential of fine-tuning in enhancing model capabilities beyond their zero-shot performance.

\textbf{Impact of Dataset Scale on Model Performance.} To assess the performance and scalability of LLMs, we also benchmark them using the \revqa at multiple size increments. By gradually increasing the number of questions and images, we aim to observe how each model's accuracy relative to the amount of data. 
This approach will allow us to identify any limitations in model capacity, potential bottlenecks, and the overall robustness of each model under different data volumes.
The results and visualizations are presented in Section 3 of the supplementary material.

\subsection{Metrics}
We report the accuracy of the LLMs to assess their performance. In open-ended scenario, where the LLM outputs are in free-text format, comparing responses directly with the ground truth answers can be challenging. To address this, we tested several string-matching methods to compare the output with the ground truth answer. We employ three approaches - \emph{exact match}, \emph{substring}, and \emph{semantic similarity} - which offer different strategies for aligning and scoring the outputs based on varying degrees of match tolerance and answer format flexibility.
\hl{The \emph{exact match score} is 1 if the answer provided by a model precisely matches the ground truth, and 0 otherwise. However, since foundation models often provide free-text answers, exact match may not accurately reflect performance. The \emph{substring score} is defined as 1 if the answer provided by the model is contained in the list of words from the ground truth answer, and $0$ otherwise.}
Nevertheless, this approach falls short in many cases; for example, when the ground truth is "male" and the prediction is "a man," the substring score is $0$, which is incorrect. Therefore, the \emph{semantic similarity score} is necessary to capture the semantic quality of the generated answer and provide a more meaningful evaluation of the LLMs. The score is based on the similarity of the generated answer and the ground truth computed using a widely recognized similarity all-MiniLM-L6-v2~\cite{reimers-2019-sentence-bert}, which is a compact, efficient model for semantic textual similarity tasks.
The reported scores are the average values over the entire dataset.
Section 3 of the supplementary material provides a more comprehensive report of the inference results.

\begin{table}[tb]
    \centering
    \small
    \begin{tabular}{@{}l|cc|cccc@{}}
        \textbf{Model} & \textbf{R-U} &  \textbf{R-B} &\textbf{OK-VQA} & \textbf{VQAv2} & \textbf{GQA} \\
        
        \midrule
        
        BLIP-2 & 46.4 & 46.1 & 45.9 & 65.0 & \textbf{44.7} \\
        InstructBLIP & \textbf{43.7} & 43.9 & 66.3 & 69.8 & 47.9 \\
        mPLUG-Owl2 & \textbf{22.1} & 22.2 & 57.7 & 79.4 & 56.1 \\
        Idefics2 & 50.8 & \textbf{49.7} & 54.6 & 70.3 & - \\
        Mantis-SigLIP & 37.6 & \textbf{37.2} & 55.4 & 74.9 & - \\
        Mantis-Idefics2 & 43.7 & \textbf{42.9} & 52.6 & 77.6 & - \\
        mPLUG-Owl3 & 47.9 & \textbf{44.8} & 60.1 & 82.1 & 65.0 \\
        GPT-4o & 62.8 & \textbf{60.8} & 71.8 & - \\
        LLaVA-OV & \textbf{50.5} & 50.6 & 66.9 & 81.1 & - \\
        Qwen2.5-VL & 59.3 & \textbf{58.1} & 84.9 & 88.6 & - \\
        PaliGemma-2 & 40.0 & \textbf{39.8} & 63.6 & 71.8 & - \\
        PaliGemma-2-Mix & 43.7 & \textbf{41.8} & 86.8 & 90.4 & - \\
        SmolVLM-Instruct & 34.5 & \textbf{31.7} & 49.1 & 35.4 & - \\

        \midrule

        Mantis-SigLIP & 67.8 & \textbf{67.6} & 99.2 & 95.5 & - \\
        Mantis-Idefics2 & 68.7 & \textbf{68.5} & 98.9 & 97.3 & - \\
        mPLUG-Owl3 & 68.9 & \textbf{68.1} & 99.1 & 97.8 & - \\
        GPT-4o & 76.6 & \textbf{73.4} & 96.7 & - & - \\
        LLaVA-OV & 56.8 & \textbf{52.4} & 94.6 & 96.1 & - \\
        PaliGemma-2-Mix & 65.6 & \textbf{64.7} & 95.8 & 97.8 & - \\
        
        \bottomrule
    \end{tabular}
    \caption[]{The benchmark scores across various models on our datasets, compared to those on previous VQA datasets. The top section presents scores in the open-ended scenario, while the bottom section shows scores in the multiple-choice scenario. The accuracies are computed using the \emph{semantic similarity} string-matching method. We emphasize the lowest scores for each row. "R-U" and "R-B" refer to ReasonVQA-U and ReasonVQA-B.} 
    \label{table:benchmark-scores}
    \vspace{-0.25cm}
\end{table}

\imgbig{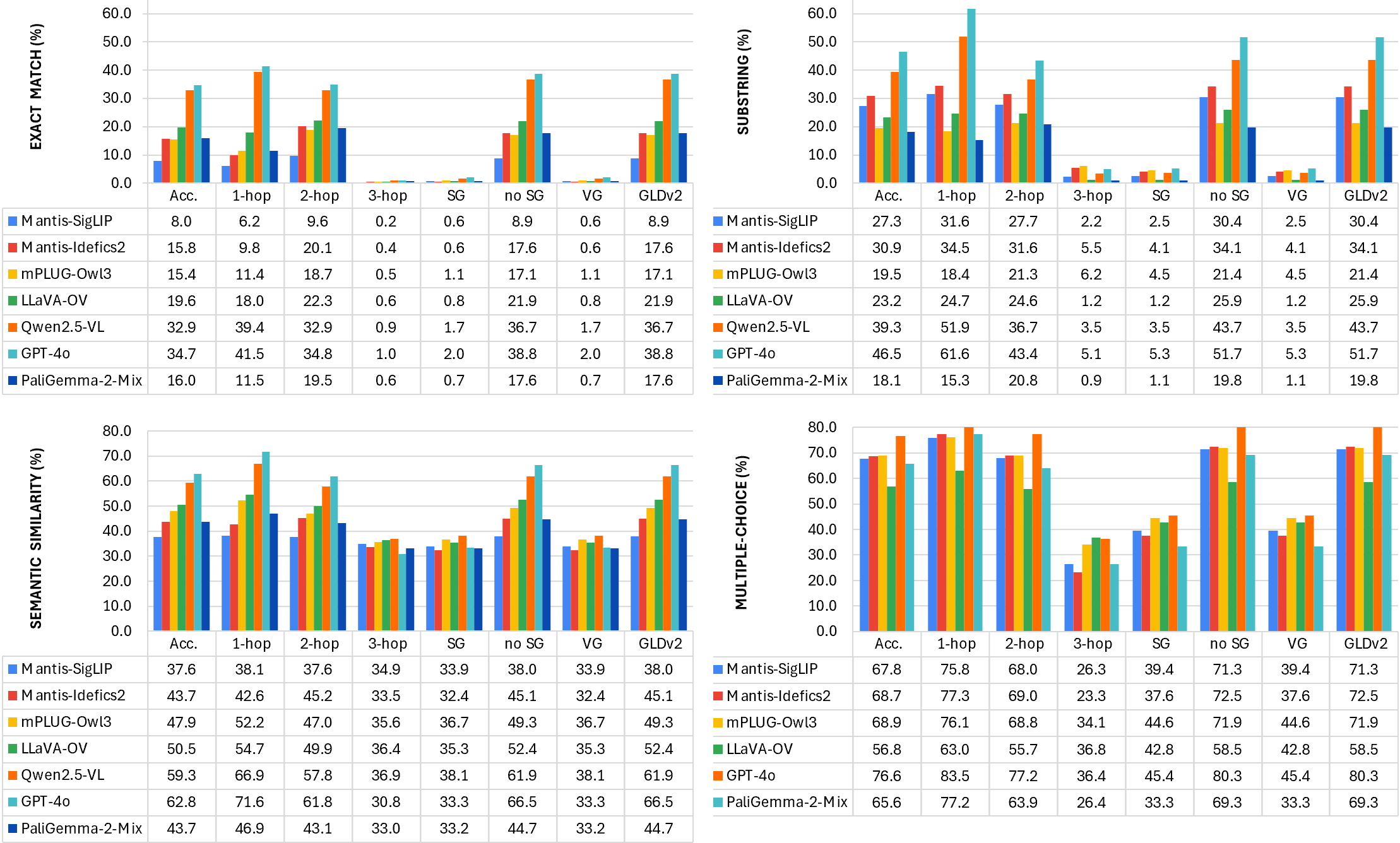}{The evaluation results in four string-matching strategies by the number of hops, by the availability of scene graph, and by the source CV dataset. "Acc." refers to the overall accuracy. "SG" refers to scene graph. "VG" and "GLDv2" refer to Visual Genome and Google Landmarks Datasets v2 respectively.}{img-benchmark-breakdown}

\subsection{Experimental Results}
Table \ref{table:benchmark-scores} presents the accuracies of tested models on two versions (balanced and unbalanced) of our dataset and on three datasets: VQAv2, OK-VQA, and GQA.
We observed that in most experiments, the scores decreased, sometimes significantly, when evaluating models on our datasets compared to their performance on previous VQA datasets. In addition, the accuracies on our balanced version are slightly lower to those on the unbalanced version.

In the zero-shot open-ended scenario, older models like BLIP-2, InstructBLIP, and mPLUG-Owl2 show moderate performance, with a notable drop on \revqa compared to VQAv2 and OK-VQA and a slight decrease relative to GQA. mPLUG-Owl2, in particular, struggles significantly, indicating a substantial gap between the model’s general VQA capabilities and its ability to handle the complexities of our datasets. Moving to more recent models, GPT-4o, Qwen2.5-VL, and PaliGemma-2-Mix emerge as the top-performing approaches. GPT-4o demonstrates its advanced reasoning and multimodal capabilities by achieving the highest accuracy. Qwen2.5-VL also performs strongly, surpassing other models across benchmarks. PaliGemma-2-Mix performs exceptionally well on previous datasets but sees a sharp drop on \revqa, suggesting that even state-of-the-art models may not generalize effectively to our datasets. SmolVLM-Instruct, as the most lightweight model among the recent ones, exhibits the lowest performance across all datasets. Its accuracy on our dataset is particularly low, highlighting its limited capability in handling the complexity of our benchmark.
On the other hand, all tested models were able to achieve remarkably higher accuracies in the zero-shot multiple choice scenario. Nevertheless, the scores on \revqa experienced a significant drop compared to those on other datasets.

For recent models, we also reported in Figure~\ref{fig:img-benchmark-breakdown} the accuracies in four string-matching approaches across different categories: the number of hop in the questions (1-hop, 2-hop, 3-hop); the availability of scene graph and the image source. 
Our results revealed that exact matching yields low overall accuracies because these models treat questions as open-ended, generating varied, free-text responses that, while often not entirely incorrect, lack the specificity required for exact matches (e.g., "very high" instead of "390 meters").
The substring approach, which considers all words from both the predicted answers and the ground truth, yields more positive scores. 
Semantic similarity matching consistently achieves higher scores, as it considers synonymous expressions, offering a more adaptable evaluation that closely aligns with human judgment.

Regarding the number of hops in questions, it is consistent across all tested models that they perform much worse on 3-hop questions compared to 1-hop and 2-hop questions. This confirms the high complexity of 3-hop questions and the need for proper reasoning steps to answer them accurately. However, it is worth noting that the performance of most models on 1-hop questions is only slightly better than on 2-hop questions, and in some cases, even worse. One possible reason for this is that while 2-hop questions are indeed more complex, they are generally longer and provide more contextual information. Consequently, certain models are able to effectively utilize the additional context in 2-hop questions, resulting in better performance. Figure~\ref{fig:img-benchmark-breakdown} also presents scores for two types of questions: those generated with information from scene graphs and those generated without such information. Generally, the scores for questions with scene graph information are lower than the other. This suggests that incorporating scene graphs to generate questions indeed increases the complexity of ReasonVQA. This valuable insight helps to guide future efforts to further increase the complexity of the dataset.

\begin{table}[tb]
    \centering
    \small
    \begin{tabular}{@{}l|ccc@{}}
        \textbf{Model} & \textbf{Zero-shot Acc.} & \textbf{Fine-tuned Acc.}\\        
        \midrule
        Qwen2-VL-7B-Instruct & 59.0 & 65.0 (+10.1\%) \\
        PaliGemma-2-3B-Mix & 40.1 & 66.8 (+66.5\%) \\        
        PaliGemma-2-10B-Mix & 65.6 & 74.5 (+13.5\%) \\
        \bottomrule
    \end{tabular}
    \caption[]{Fine-tuning performance of three LLMs on our dataset. The table compares accuracy before and after fine-tuning, highlighting the improvement achieved through training. Improvement percentages are shown in parentheses.} 
    \label{table:finetuning}
    \vspace{-0.25cm}
\end{table}

For the fine-tuning experiment, we fine-tune three vision-language models, Qwen2-VL-Instruct 7B~\cite{DBLP:journals/corr/abs-2409-12191}, PaliGemma-2-Mix 3B, and PaliGemma-2-Mix 10B~\cite{DBLP:journals/corr/abs-2412-03555,PaliGemma2Mix2025}, on our dataset to evaluate its effectiveness in adapting pre-trained models to knowledge-based VQA. Fine-tuning is conducted using LoRA~\cite{DBLP:conf/iclr/HuSWALWWC22} to reduce computational overhead while preserving model capacity. For each model, we compare performance before and after fine-tuning using \revqa train set. Table \ref{table:finetuning} presents the results on the test set, showing a significant improvement in accuracy across these models, demonstrating the dataset's effectiveness in knowledge-based reasoning. Notably, the PaliGemma-2-3B-Mix achieves significant gains with lower resource, suggesting that smaller models have more room for enhancement when adapted to a specific dataset.


\section{Conclusions}
\label{sec:conclusions}
\hl{We have introduced a novel, cost-effective, and scalable VQA dataset that requires external knowledge for answering questions, demanding minimal manual effort. Our construction framework leverages existing CV dataset annotations and external knowledge base Wikidata. While currently built from specific image sources, the framework is generic, allowing expansion to integrate images from other CV datasets and maintain question diversity. Users can also tailor the dataset to specific interests by selecting from 20 predefined knowledge domains. ReasonVQA features complex questions derived from scene and multi-hop knowledge graphs, proving highly challenging for state-of-the-art VQA models.}
We provide source code for easy integration of new images and knowledge sources, enabling researchers to expand and customize their own datasets. We believe this resource will be valuable for evaluating existing models and advancing more robust VQA systems.

\section*{Acknowledgements}
This work was partially funded by the European Union’s programme under grant agreement No.101092908 (SMARTEDGE), by the Chips Joint Undertaking (JU), European Union (EU) HORIZON-JU-IA, under grant agreement No. 101140087 (SMARTY) and by the German Research Foundation (DFG) under the
COSMO project (ref. 453130567).

{
    \small
    \bibliographystyle{ieeenat_fullname}
    \bibliography{main}
}

\clearpage

\appendix

\section*{Appendices}

\addcontentsline{toc}{section}{Appendices}

\setcounter{section}{0}


\renewcommand{\thesection}{\arabic{section}}

\setcounter{section}{0}
\setcounter{figure}{0}
\setcounter{table}{0}
\setcounter{equation}{0}
\setcounter{page}{1} 

\section{The ReasonVQA Framework Additional Details}
This is additional content for Section 3 in the main paper. Here, we provide more details regarding the question generation process, the integration of external knowledge, and the visualization of answer distribution balancing and dataset splitting.

\subsection{Template-based Question Generation}
Our framework consists of three steps: (1) External Knowledge Integration, (2) Question Generation, and (3) Dataset Construction. In addition to Figure 2 in the main paper, Algorithm \ref{alg:allsteps} also describes the detailed workflow of these steps.

\begin{algorithm}[ht!]
\caption{Algorithm for generating questions and answers from annotated images}\label{alg:allsteps}
\small
\DontPrintSemicolon%
\SetKwInOut{Input}{Input}%
\SetKwInOut{Output}{Output}%
\SetKwFor{Loop}{Loop}{}{EndLoop}
\newcommand\mycommfont[1]{\footnotesize\ttfamily\textcolor{gray}{#1}}%
\SetCommentSty{mycommfont}%
\Input{Annotated image \textbf{Img}}%
\Output{A set of questions \textbf{Q} generated for \textbf{Img}} %

\BlankLine
\tcc{Step 1: External Knowledge Integration (Main Paper Section 3.1)}%
$\{\mathcal{C}_i\}$ = set of Wikidata entities corresponding to annotated objects in \textbf{Img}\;%
$\mathcal{G}_i (\mathcal{V}_i, \mathcal{E}_i)$ = knowledge graph from Wikidata with root $\mathcal{C}_i$ \;%
$\mathcal{E}$ = $\{\mathcal{E}_i\}$  \tcp{set of potential properties}

\BlankLine
\tcc{Step 2: Question Generation (Main Paper Section 3.2)}%
$\mathcal{T}_m$ = set of main templates $\forall e_i \in \mathcal{E}$ \;%
$\mathcal{T}_s$ = set of sub-clause templates $\forall e_i \in \mathcal{E}$ \;%

\SetKwFunction{FMain}{Generate}
\SetKwProg{Fn}{Function}{:}{}
\Fn{\FMain{$e_j$}}{
    \eIf{$ j = 0 $} {
        $t \gets \mathcal{T}_m[e_j]$ \;
    }{
        $t \gets \mathcal{T}_s[e_j]$ \;
    }
    \Return $t \cup ~\FMain{$e_{j+1}$}$ \;
}

\BlankLine
$\mathcal{D}_1$ = empty dataset \;
\ForEach{$v \in \mathcal{V}_i$} {
    $\{e_j\}$ = set of edges from $v$ to $\mathcal{C}_i$ \;%
    $\mathcal{D}_1$ = $\mathcal{D}_1$ $\cup$ ~\FMain{$e_0$}
}

\BlankLine
\tcc{Step 3: Dataset Construction (Main Paper Section 3.3)}
$\mathcal{D}_2=balance(\mathcal{D}_1)$ \tcp{balance the answer distribution}
$\mathcal{D}_3=split(\mathcal{D}_2)$ \tcp{split into train set and test set}
\end{algorithm}

\subsection{Concept Linking and Template Construction}
\textbf{Concept Linking between Image and Knowledge Base.}
The process of linking an annotated object in an image to a concept in a knowledge base may vary depending on the computer vision (CV) dataset and the knowledge base (KB) used. 

For Visual Genome (VG)~\cite{DBLP:journals/ijcv/KrishnaZGJHKCKL17}, we leverage the WordNet~\cite{DBLP:journals/cacm/Miller95} synset names provided in the annotations to identify the corresponding entity in Wikidata. Specifically, for each object associated with a synset name, we convert the synset name into a synset ID using NLTK. We then query the respective Wikidata entity via SPARQL. Figure \ref{fig:img-linking-vg} shows an example of linking the object \emph{traffic light} in the image to the corresponding concept with the same name in Wikidata. Since the bounding box of the traffic light was annotated with the synset name \texttt{"traffic\_light.n.01"}, we convert it into synset ID \texttt{"06887235-n"} using the NLTK~\cite{DBLP:books/daglib/0022921} package and then search for the Wikidata entity associated with this synset ID via SPARQL.

\img{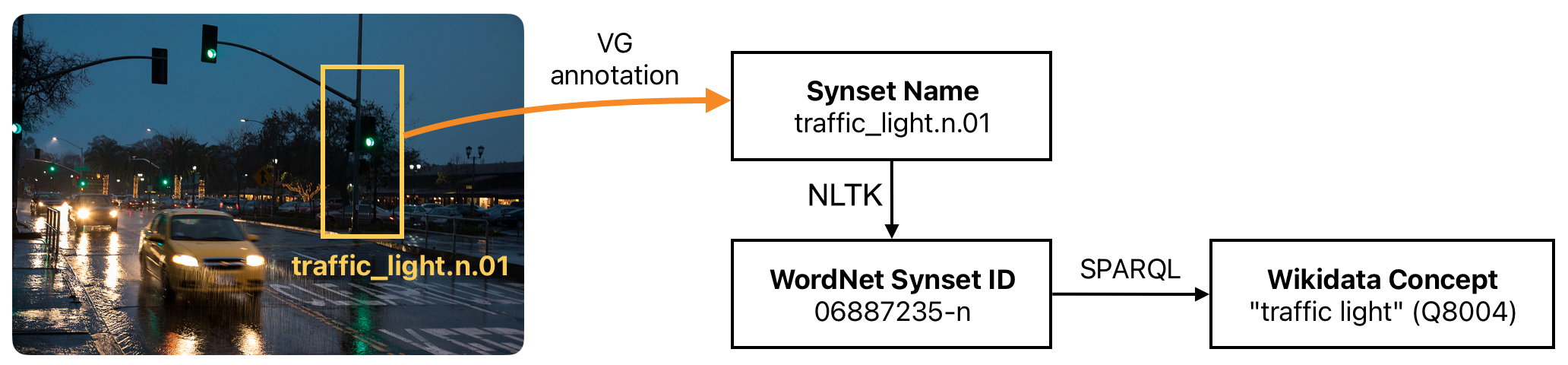}{Example of linking an object from VG to a concept in Wikidata using Wordnet synset name. The Wikidata entity is retrieved by the WordNet synset ID, which is converted from the synset name using the NLTK package.}{img-linking-vg}

For Google Landmarks Dataset v2 (GLDv2)~\cite{DBLP:conf/cvpr/WeyandACS20}, from the Wikimedia URLs provided in the annotations, we heuristically extract the name of the landmark. Then we search for the Wikidata concept by this name. In Figure~\ref{fig:img-linking-gld}, we extract the name \emph{Maria Magdalena kyrka, Stockholm} from the Wikimedia URL. With a simple SPARQL query, we can search for the entity that links to the Wikimedia Commons resource with this name.

\img{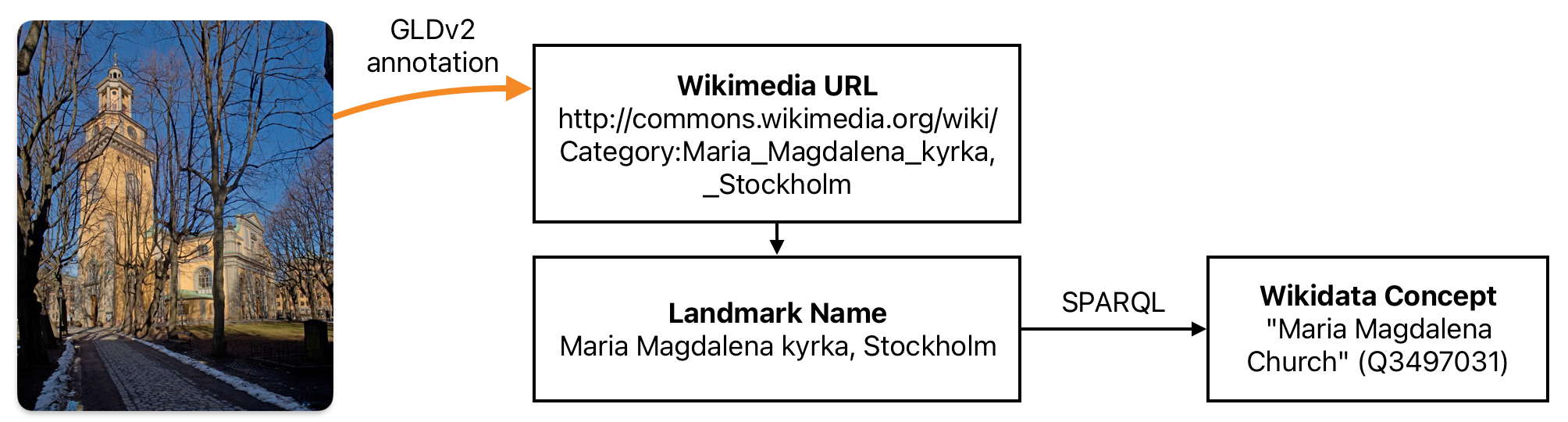}{Example of linking a landmark from GLDv2 to a concept in Wikidata. The Wikidata entity is retrieved by its name, which is extracted from the Wikimedia URL in GLDv2.}{img-linking-gld}

\textbf{Main Template and Sub-clause Template Crafting.}
After connecting an object in the image to an entity in the KB, referred to as the \emph{root concept}, we begin gathering multi-hop knowledge. Initially, we retrieve knowledge around the root concept in the form of triplets. Each triplet corresponds to a property and connects the root concept to either a literal value or another concept. If the end of a triplet is another concept, we continue gathering knowledge for this one. This process of traversing through the knowledge graph yields multi-hop knowledge. In practice, we find that traversing up to three hops strikes a balance between complexity and minimizing grammatical errors in generated questions.

During the process of fetching knowledge from the KB, we also collect all potential properties and manually created templates for them, then add them to our \emph{template bank}. It is important to note that for each property, we only need to predefine templates the first time our system encounters this property. For instance, if a concept has the property "country", we define templates for this property just once and add to the bank. Subsequent concepts with the same property can then reuse these templates from the bank. That means the number of templates to be hand-crafted will gradually decrease until all potential properties have corresponding templates in our bank, at which point the question generation process becomes completely automatic.

Specifically, for each property, we define a \emph{main template} and an optional \emph{sub-clause template}. Our template bank consists of 182 main templates and 100 sub-clause templates for 182 distinct properties. These numbers can increase as our framework can be extended to include additional image sources and knowledge bases. Table~\ref{table:ds-templates} presents a few examples in our template bank.

\begin{table}[ht!]
\centering
\begin{tabular}{ @{}cl@{} }
\toprule
\textbf{Property} & \textbf{Templates} \\
\midrule
\multirow{2}{*}{architect}  & (a) \texttt{Who designed \_\_ ?} \\
& (b) \texttt{the architect of \_\_} \\
\midrule
\multirow{2}{*}{author}  & (a) \texttt{Who created \_\_ ?} \\
& (b) \texttt{the author of \_\_} \\
\midrule
\multirow{2}{*}{country} & (a) \texttt{In which country is \_\_ located?} \\
& (b) \texttt{where \_\_ is located} \\
\midrule
height & (a) \texttt{How high is \_\_?} \\
\midrule
width & (a) \texttt{How wide is \_\_?} \\
\midrule
\multirow{2}{*}{official language} & (a) \texttt{What is the official language of \_\_?} \\
& (b) \texttt{the official language of \_\_} \\
\midrule
\multirow{2}{*}{currency} & (a) \texttt{What is the currency of \_\_?} \\
& (b) \texttt{the currency of \_\_} \\
\midrule
\multirow{2}{*}{capital} & (a) \texttt{What is the capital of \_\_?} \\
& (b) \texttt{the capital of \_\_} \\
\midrule
\multirow{2}{*}{mother} & (a) \texttt{Who is the mother of \_\_?} \\
& (b) \texttt{the mother of \_\_} \\
\midrule
\multirow{2}{*}{place of birth} & (a) \texttt{Where was \_\_ born?} \\
& (b) \texttt{the place of birth of \_\_} \\
\bottomrule
\end{tabular}
\caption{Examples of predefined templates. For each property, we define a (a) \emph{main template} and an optional (b) \emph{sub-clause template}.}
\label{table:ds-templates}

\end{table}

\subsection{Answer Distribution Balancing}
\label{sec:balancing}
To reduce the bias in the answer distribution, we iteratively apply a balancing process following three criteria: (1) preserving the relative size of \emph{head} and \emph{tail}; (2) maintaining the frequency order; and (3) prioritizing the removal of answers associated with a higher number of questions. The \emph{head} represents the group of questions with the most answers, while the \emph{tail} represents the group with the least. Figure \ref{fig:img-balancing} illustrates an example of the answer distribution before and after applying the balancing process for 10 and 20 rounds. Questions and answers are distributed into groups based on the properties from which they were generated. While the balancing process applies to all groups, we visualize only the top 20 groups with the highest number of answers. For each group, we also visualize only the top 10 most frequent answers, in descending order. After 20 iterations, 26,100 questions were discarded, which is 33.4\% of the total number of questions. The answer distribution became much more balanced, with a few groups on the left side showing the most noticeable improvement.

\imgbig{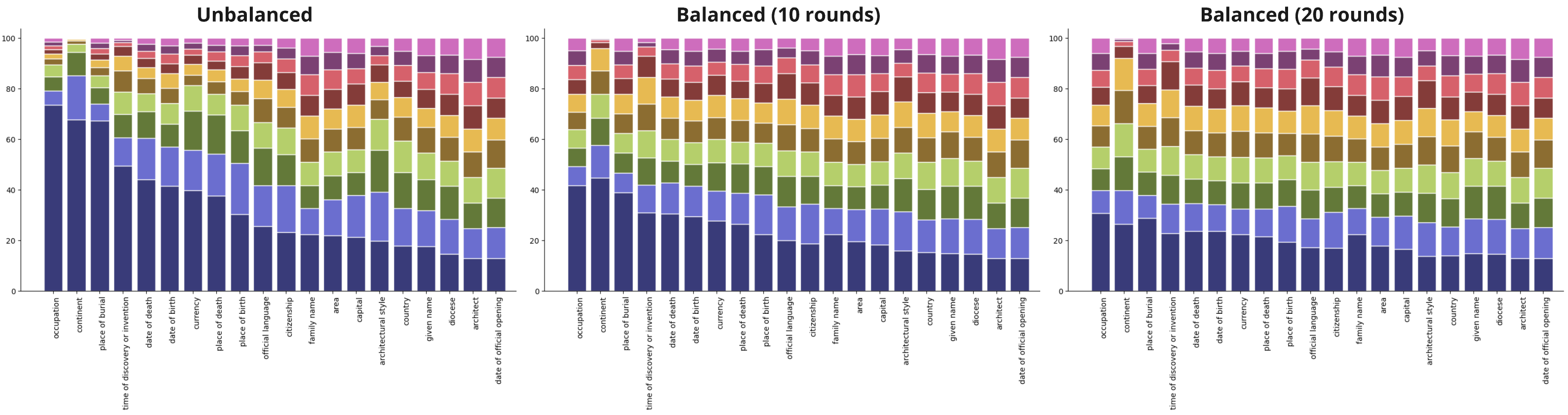}{The answer distribution for the unbalanced dataset (left) and the balanced datasets after 10 rounds (middle) and 20 rounds (right) of the balancing process. We show the top 10 answers in the top 20 groups. The column height corresponds to the relative frequency of each answer. The distribution started from being heavily biased to becoming more uniform.}{img-balancing}

\subsection{Dataset Splitting}
Figure \ref{fig:img-split} shows the similarity in answer distribution between the train set and test set. Here we also visualize top 10 most frequent answers from the top 20 groups.

\img{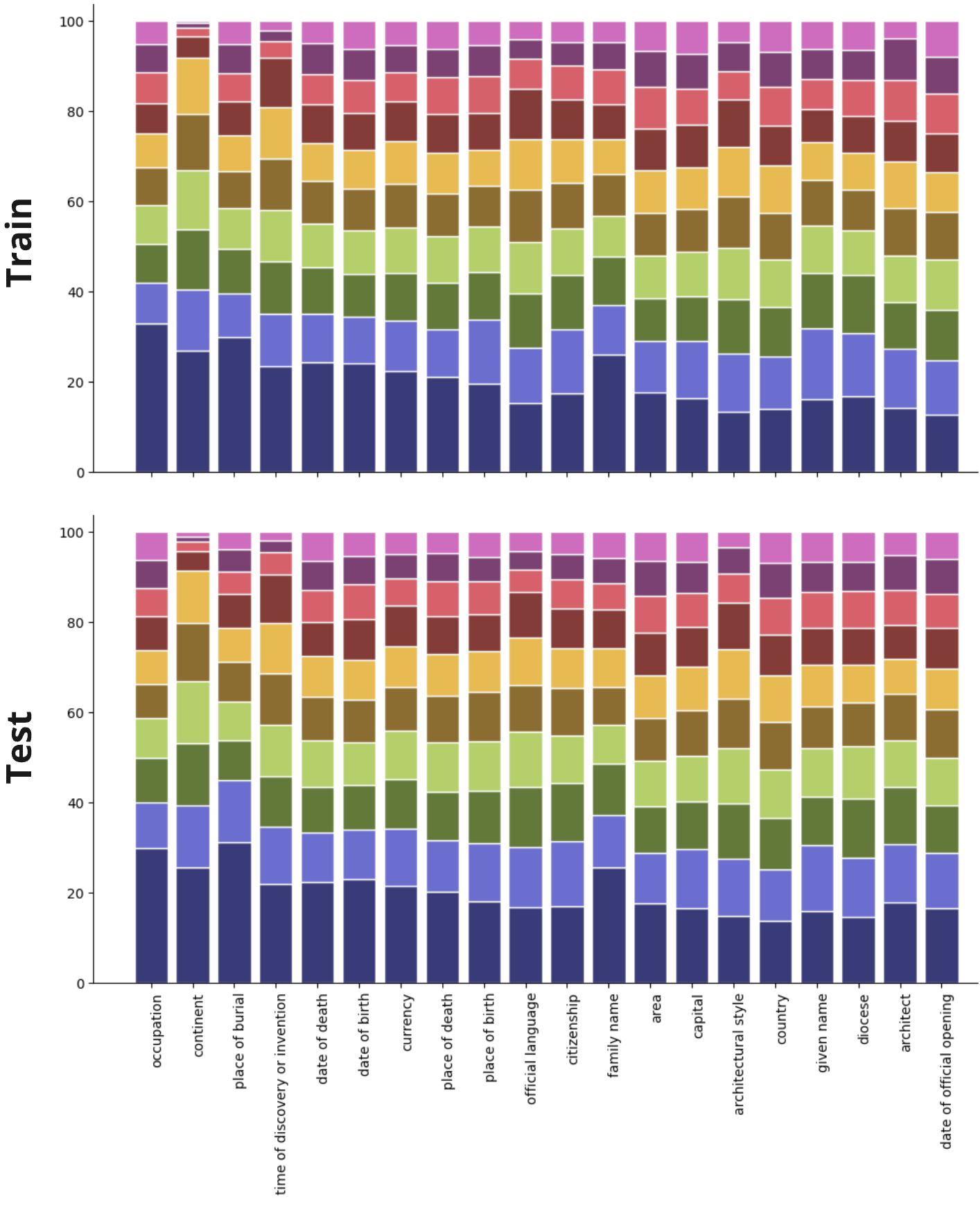}{The answer distribution of the train and test set. The same top 10 answers in the top 20 groups for the train set (top) and test set (bottom). The column height corresponds to the relative frequency of each answer.}{img-split}

\vspace{0.1cm}
\section{Dataset Analysis}
In this section, we present more statistics of our dataset and provide details of our user study conducted for question quality evaluation.

\subsection{Dataset Statistics and Examples}
The latest version of ReasonVQA consists of nearly 4.2M generated from 598K images, with 1.3M 1-hop questions, 2.8M 2-hop questions, and 5.4K 3-hop questions. Our dataset statistics are shown in Table~\ref{table:ds-stats}. Figure~\ref{fig:img-word-count} illustrates the distribution of questions by the first four words. Figure~\ref{fig:img-demo} presents multiple instances from ReasonVQA.

\begin{table}[ht]
\centering
\begin{tabular}{@{}m{0.35\linewidth} c c c@{}}
\toprule
& \textbf{ReasonVQA} & \textbf{ReasonVQA-U} \\
\midrule
\# Images & 598,525 & 13,326 \\
\# Questions & 4,174,024 & 78,007 \\
\# 1-hop questions & 1,358,634 & 23,767 \\
\# 2-hop questions & 2,809,960 & 49,459 \\
\# 3-hop questions & 5,430 & 4,781 \\
\# Unique questions & 123,204 & 22,368 \\
\# Unique answers & 73,068 & 9,103 \\
\# Unique choices & 123,411 & 21,037 \\
Avg. question length (words) & 9.77 & 9.62 \\
Avg. answer length (words) & 1.53 & 1.49 \\
\bottomrule
\end{tabular}
\caption{Some characteristics of our datasets in full version and subset version.}
\label{table:ds-stats}
\end{table}

\img{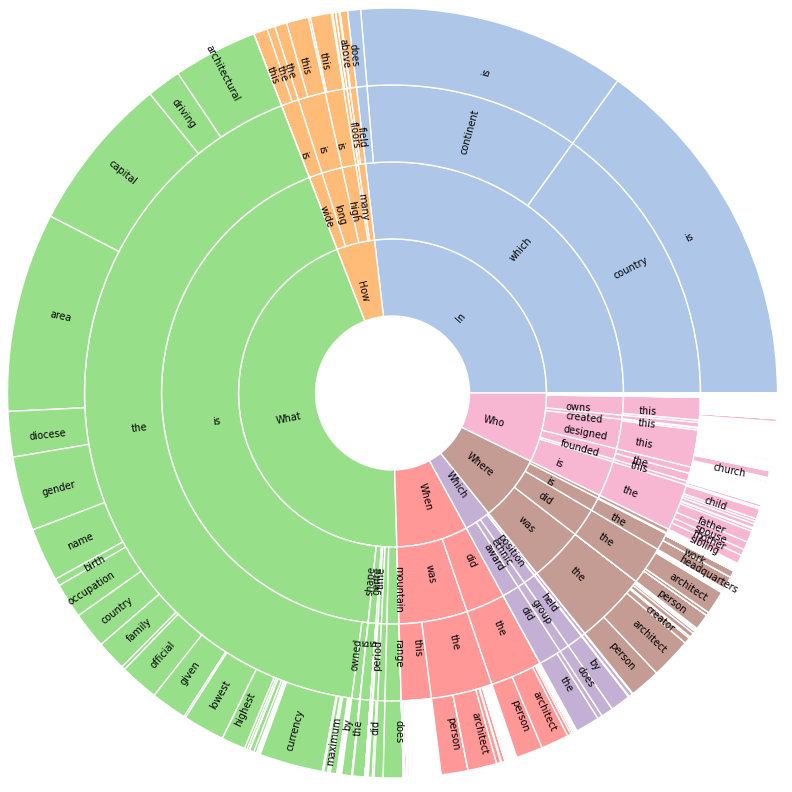}{Distribution of questions by first four words}{img-word-count}

\imgbig{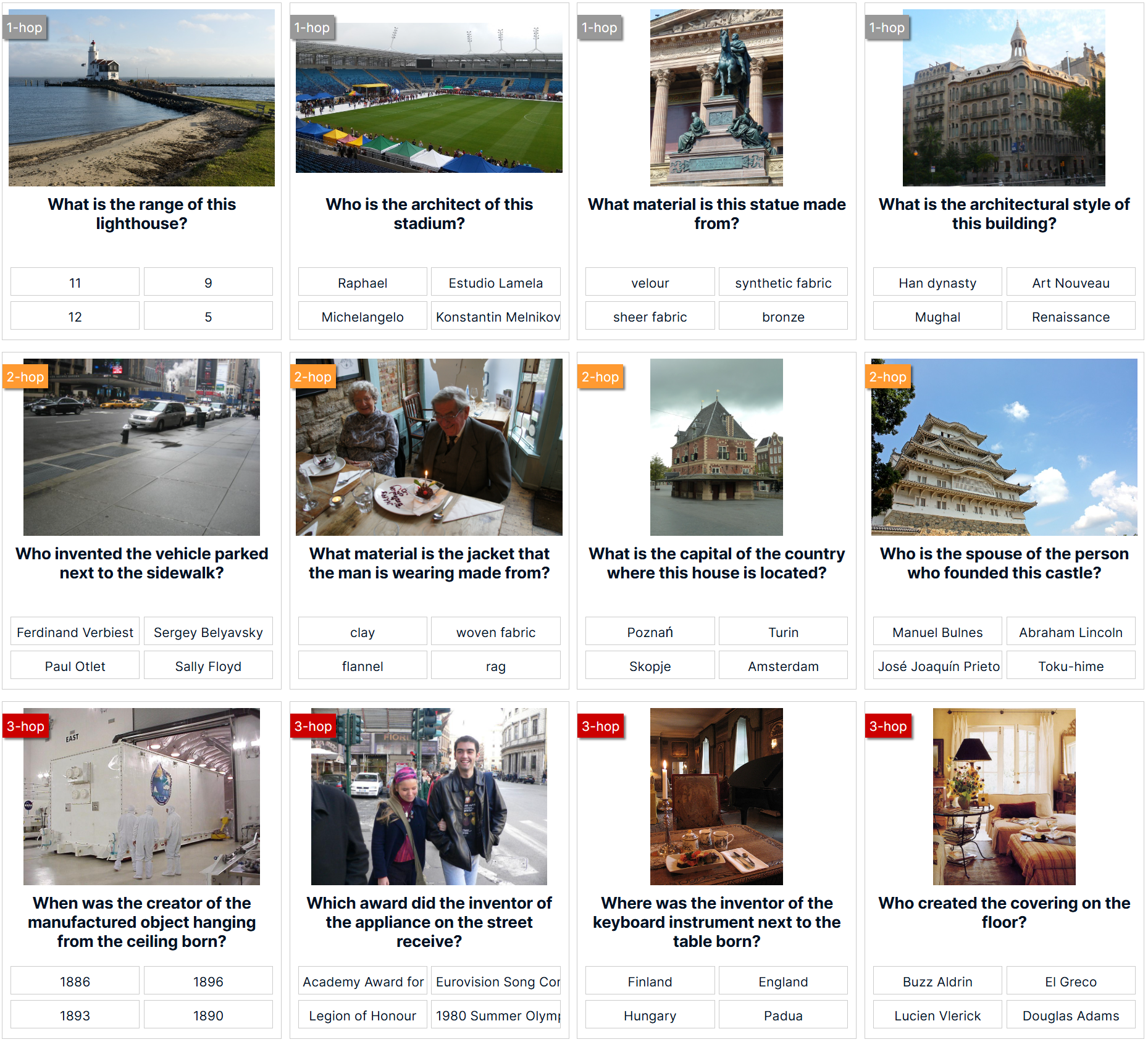}{Some example questions and answers from ReasonVQA. The first row shows 1-hop questions, the middle row shows 2-hop questions with the first two questions constructed by incorporating the scene graph, and the last row contains 3-hop questions.}{img-demo}

\subsection{Dataset Domains}
As detailed in the main paper, we categorized questions into 20 domains, outlined as follows. 
Figure~\ref{fig:img-breakdown} visualizes the domain distribution.

\begin{enumerate}
    \item \textbf{Places \& Locations }\\e.g. Country where a place is located
\item \textbf{Person \& Institutions }\\e.g. Organization employing an individual
\item \textbf{Temporal Concepts }\\e.g. Official opening date
\item \textbf{Characteristics \& Properties }\\e.g. Height of buildings or structures
\item \textbf{Language \& Cultural }\\e.g. Language officially recognized in a region
\item \textbf{History \& Events }\\e.g. Date or people associated with a historical event
\item \textbf{Physical Geography }\\e.g. Capital city of a country
\item \textbf{Politics \& Ideologies }\\e.g. Head of government
\item \textbf{Economics \& Labor }\\e.g. Industry associated with an organization
\item \textbf{Nature \& Human Interaction }\\e.g. Water composition of a given area
\item \textbf{Technology \& Innovation }\\e.g. Manufacturer of a technological item
\item \textbf{Science \& Quantitative Analysis }\\e.g. Temperature or light range of an object
\item \textbf{Health \& Medicine }\\e.g. Symptoms associated with a condition
\item \textbf{Education \& Knowledge Systems }\\e.g. Institution where an individual was educated
\item \textbf{Art \& Creative Expressions }\\e.g. Collection housing an artistic work
\item \textbf{Philosophy \& Spiritual Beliefs }\\e.g. Entity or concept to which a church is dedicated
\item \textbf{Media \& Communication Systems }\\e.g. Number of episodes in a series
\item \textbf{Environment \& Sustainability }\\e.g. Inflow and outflow of lakes
\item \textbf{Law \& Justice Systems }\\e.g. Area of legal authority
\item \textbf{Food \& Nutrition }\\e.g. Caloric content of food or drink
\end{enumerate}

\imgfix{images/breakdown_category.png}{Distribution of questions across domains. Each question is categorized into specific domains based on the property from which it was generated.}{img-breakdown}

\subsection{Question Evaluation by User Study}

\imgfix{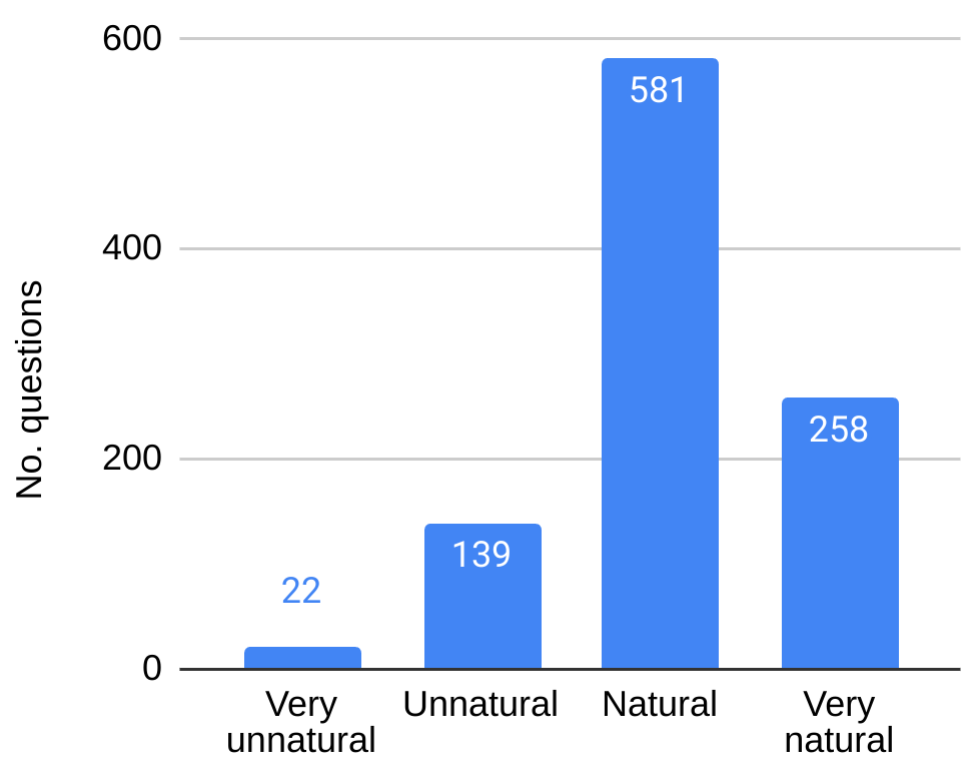}{Visualization of the user study for question naturalness evaluation.}{img-user-study}

To evaluate the quality of generated questions, we conducted a user study with 1,000 randomly selected question and image pairs. Twenty participants, all proficient in English as their primary language for work or study, assessed the correctness of the answers and the naturalness of 50 randomly chosen questions each. For the naturalness, they rated the questions on a four-level scale: (1) very unnatural, (2) unnatural, (3) natural, and (4) very natural. Additionally, they were also asked to mark any questions with grammatical errors. The results indicated that 96\% of selected answers are correct, 2.2\% of the questions were rated as "very unnatural," 13.9\% as "unnatural," 58.1\% as "natural," and 25.8\% as "very natural,", with only 2.5\% of the questions had grammatical errors. The results are shown in Figure \ref{fig:img-user-study}.

\section{Experiments}
In this section, we provide experimental details regarding different dataset sizes. We also present the average accuracies across domains, along with additional metrics for the benchmarked models, including the Standard Deviation (SD) and the Standard Error of the Mean (SEM). These scores are further analyzed across various aspects of our dataset.

\subsection{Benchmark Results Across Dataset Sizes}
To understand how models perform with varying dataset sizes, we conducted experiments on seven models
Mantis-SigLIP~\cite{DBLP:journals/tmlr/JiangHZWKLC24},
Mantis-Idefics2~\cite{DBLP:journals/tmlr/JiangHZWKLC24},
mPLUG-Owl3~\cite{DBLP:journals/corr/abs-2408-04840},
GPT-4o~\cite{DBLP:journals/corr/abs-2410-21276},
LLaVA-OV~\cite{DBLP:journals/tmlr/0080ZGZ00ZZL0L25},
Qwen2.5-VL~\cite{DBLP:journals/corr/abs-2502-13923},
PaliGemma-2-Mix~\cite{PaliGemma2Mix2025}
across a range of dataset sizes.
In our experiments, we aimed to investigate how model performance varies with dataset size while controlling for sample difficulty. To this end, we first divided the full dataset into three pools based on the number of hops. For each target dataset size (e.g., $10k$, $20k$, …, $80k$ samples), we randomly sampled examples from each pool proportionally, maintaining the same ratio of as in the full dataset. This strategy ensures that each subset reflects the overall difficulty distribution, enabling us to evaluate how model performance scales with dataset size without bias toward easier or harder samples.
Figure~\ref{fig:img-ds-sizes} illustrates the performance of various models across different dataset sizes. We observe that as the dataset grows, model performance initially improves but then declines with varying degrees of intensity. This decline likely occurs because as new samples are introduced, they bring additional domain knowledge and more complex questions, which may challenge the ability of models to maintain efficiency and accuracy under the increased data load. This observation suggests that integrating a wider variety of image sources and knowledge domains could create a more demanding benchmark, providing a more comprehensive assessment of model robustness.

\imgfix{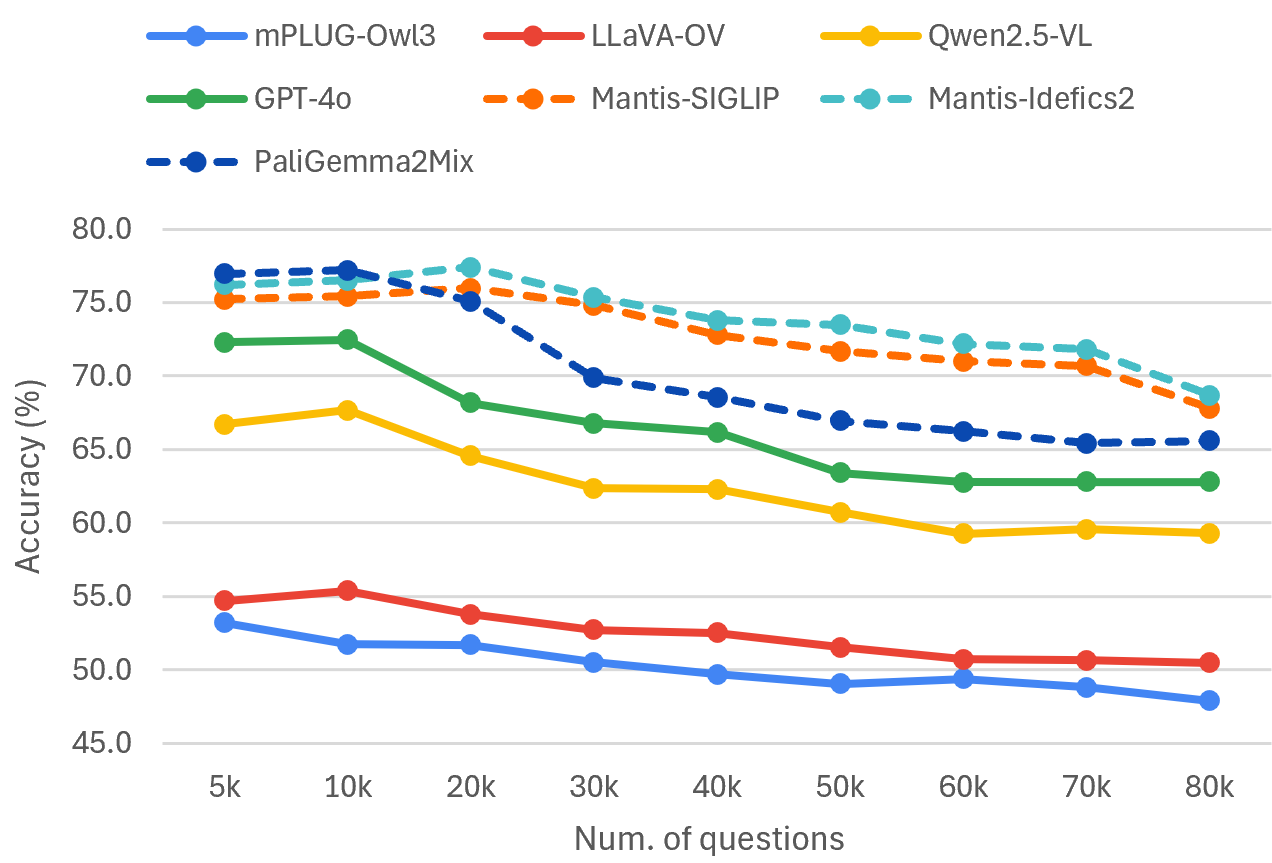}{Performance of models across different dataset sizes in open-ended (solid lines) and multiple choice (dashed lines) scenarios, with accuracy reported.}{img-ds-sizes}

\subsection{Benchmark Results Across Dataset Aspects}
For a fine-grained analysis of our dataset, we evaluated the accuracy of 13 large language models (LLMs) (BLIP-2~\cite{DBLP:conf/icml/0008LSH23},
InstructBLIP~\cite{DBLP:conf/nips/Dai0LTZW0FH23},
mPLUG-Owl2~\cite{DBLP:journals/corr/abs-2311-04257},
Idefics2~\cite{DBLP:conf/nips/LaurenconTCS24},
Mantis-SigLIP~\cite{DBLP:journals/tmlr/JiangHZWKLC24},
Mantis-Idefics2~\cite{DBLP:journals/tmlr/JiangHZWKLC24},
mPLUG-Owl3~\cite{DBLP:journals/corr/abs-2408-04840},
GPT-4o~\cite{DBLP:journals/corr/abs-2410-21276},
LLaVA-OV~\cite{DBLP:journals/tmlr/0080ZGZ00ZZL0L25},
Qwen2.5-VL~\cite{DBLP:journals/corr/abs-2502-13923},
PaliGemma-2~\cite{DBLP:journals/corr/abs-2412-03555},
PaliGemma-2-Mix~\cite{PaliGemma2Mix2025},
SmolVLM-Instruct~\cite{marafioti2025smolvlm}) across all domains, as shown in Figure~\ref{fig:benchmark-category}. We observed that Health \& Medicine proved to be the most challenging domain, achieving an average accuracy of only 20.7\%. This was closely followed by Art \& Creative Expressions, with an average accuracy of 22.5\%. The most difficult question types consistently involved numerical values, such as inquiries about the area of a city. Furthermore, we computed Standard Error of the Mean (SEM) and Standard Deviation (SD) scores to better illustrate the variability within our dataset, as presented in Table~\ref{table:std-open-ended} and~\ref{table:std-multichoice}. Overall, the SD values are relatively large, indicating a high degree of variability in the model performances across different aspects of our dataset. This variability indicates that \revqa poses significant challenges to the models, as large SD values typically reflect a high sensitivity to the diverse types of samples or tasks within the dataset. At the same time, the relatively low SEM values (ranging from 0.1\% to 0.7\%) suggest that the mean accuracy scores for each model are estimated with high precision, despite the substantial variability indicated by the SD. This highlights the presence of considerable individual sample variability, with the models demonstrating inconsistent performance across different subsets of the dataset.

\imgbigscale{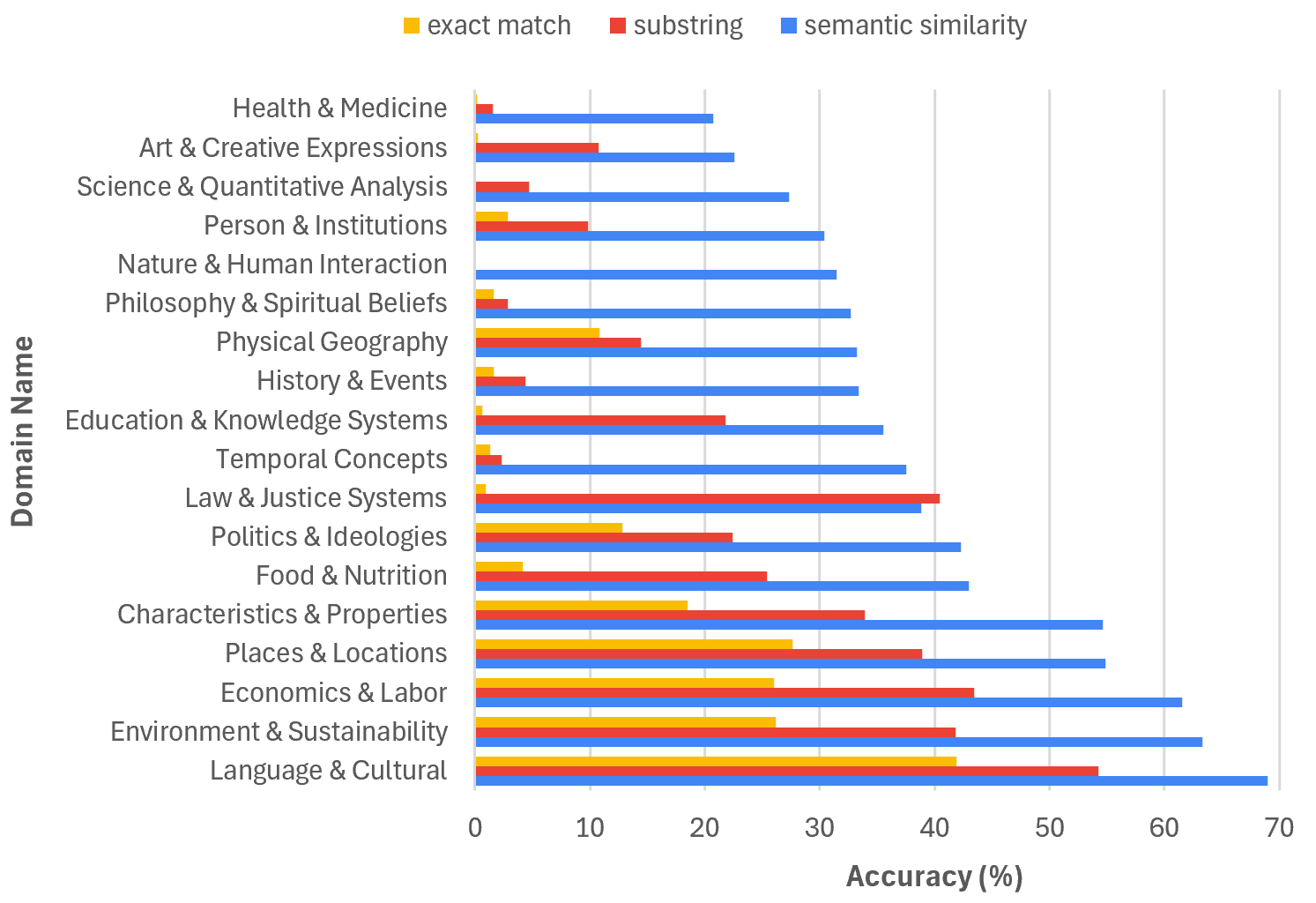}{Average accuracy of models across domains.}{benchmark-category}{0.8}

\begin{table*}
\vspace{2cm}
    \centering
    \small
    \begin{tabular}{@{}l|c|ccc|cc|cc@{}}
        \textbf{Model} & \textbf{Overall} &  \textbf{1-hop} &\textbf{2-hop} & \textbf{3-hop} & \textbf{SG} & \textbf{no SG} & \textbf{VG} & \textbf{GLDv2} \\
        
        \midrule
        
BLIP-2 & \textbf{31.3} (0.1) & \textbf{31.3} (0.2) & \textbf{31.5} (0.1) & \textbf{16.3} (0.2) & \textbf{16.6} (0.2) & \textbf{32.1} (0.1) & \textbf{16.6} (0.2) & \textbf{32.1} (0.1) \\
InstructBLIP & \textbf{30.9} (0.1) & \textbf{28.6} (0.2) & \textbf{31.7} (0.1) & \textbf{16.1} (0.2) & \textbf{18.7} (0.2) & \textbf{31.6} (0.1) & \textbf{18.7} (0.2) & \textbf{31.6} (0.1) \\
mPLUG-Owl2 & \textbf{17.2} (0.1) & \textbf{17.6} (0.1) & \textbf{17.1} (0.1) & \textbf{9.9} (0.1)& 1\textbf{0.6} (0.1)& \textbf{17.6 } (0.1)& 1\textbf{0.6} (0.1)& 1\textbf{7.6 } (0.1)\\
Idefics2 & \textbf{31.4} (0.1) & \textbf{28.5} (0.2) & \textbf{33.2} (0.1) & \textbf{18.0} (0.3) & \textbf{20.4} (0.2) & \textbf{32.0} (0.1) & \textbf{20.4} (0.2) & \textbf{32.0} (0.1) \\
Mantis-SigLIP & \textbf{26.1} (0.1) & \textbf{23.4} (0.2) & \textbf{28.0} (0.1) & \textbf{17.2} (0.2) & \textbf{18.3} (0.2) & \textbf{26.9} (0.1) & \textbf{18.3} (0.2) & \textbf{26.9} (0.1) \\
Mantis-Idefics2 & \textbf{31.0} (0.1) & \textbf{26.5} (0.2) & \textbf{33.7} (0.2) & \textbf{18.9} (0.3) & \textbf{20.5} (0.2) & \textbf{31.8} (0.1) & \textbf{20.5} (0.2) & \textbf{31.8} (0.1) \\
mPLUG-Owl3 & \textbf{29.4} (0.1) & \textbf{26.2} (0.2) & \textbf{31.3} (0.1) & \textbf{17.2} (0.2) & \textbf{20.4} (0.2) & \textbf{30.0} (0.1) & \textbf{20.4} (0.2) & \textbf{30.0} (0.1) \\

LLaVA-OV & \textbf{30.7} (0.1) & \textbf{29.2} (0.2) & \textbf{31.8} (0.1) & \textbf{18.4} (0.3) & \textbf{18.4} (0.2) & \textbf{31.3} (0.1) & \textbf{18.4} (0.2) & \textbf{31.3} (0.1) \\
Qwen2.5-VL & \textbf{33.4} (0.1) & \textbf{32.5} (0.2) & \textbf{33.8} (0.2) & \textbf{17.2} (0.2) & \textbf{20.6} (0.2) & \textbf{33.7} (0.1) & \textbf{20.6} (0.2) & \textbf{33.7} (0.1) \\
GPT-4o & \textbf{33.0} (0.1) & \textbf{31.0} (0.2) & \textbf{33.0} (0.2) & \textbf{17.8} (0.3) & \textbf{21.2} (0.2) & \textbf{32.4} (0.1) & \textbf{21.2} (0.2) & \textbf{32.4} (0.1) \\
PaliGemma-2 & \textbf{24.7} (0.1) & \textbf{22.5} (0.2) & \textbf{26.1} (0.1) & \textbf{12.1} (0.2) & \textbf{11.8} (0.1) & \textbf{25.3} (0.1) & \textbf{11.8} (0.1) & \textbf{25.3} (0.1) \\
PaliGemma-2-Mix & \textbf{31.6} (0.1) & \textbf{30.0} (0.2) & \textbf{32.9} (0.1) & \textbf{18.9} (0.3) & \textbf{19.3} (0.2) & \textbf{32.4} (0.1) & \textbf{19.3} (0.2) & \textbf{32.4} (0.1) \\
SmolVLM & \textbf{15.7} (0.1) & \textbf{15.5} (0.2) & \textbf{15.3} (0.1) & \textbf{8.7} (0.2) & \textbf{10.8} (0.2) & \textbf{15.8} (0.1) & \textbf{10.8} (0.2) & \textbf{15.8} (0.1) \\

        \bottomrule
    \end{tabular}
    \caption[]{SD and SEM scores across various models on our datasets in the zero-shot open-ended scenario. The accuracies are computed using the \emph{semantic similarity} string-matching method. SD scores are highlighted in bold, with SEM scores provided in parentheses. "SG" refers to scene graph. "VG" and "GLDv2" refer to Visual Genome and Google Landmarks Datasets v2 respectively.} 
    \label{table:std-open-ended}
\end{table*}

\begin{table*}
    \centering
    \small
    \begin{tabular}{@{}l|c|ccc|cc|cc@{}}
        \textbf{Model} & \textbf{Overall} &  \textbf{1-hop} &\textbf{2-hop} & \textbf{3-hop} & \textbf{SG} & \textbf{no SG} & \textbf{VG} & \textbf{GLDv2} \\
        
        \midrule
        
BLIP-2 & \textbf{47.9} (0.2) & \textbf{43.7} (0.3) & \textbf{48.4} (0.2) & \textbf{46.5} (0.7) & \textbf{48.9} (0.5) & \textbf{47.0} (0.2) & \textbf{48.9} (0.5) & \textbf{47.0} (0.2) \\
InstructBLIP & \textbf{47.6} (0.2) & \textbf{44.7} (0.3) & \textbf{47.4} (0.2) & \textbf{43.8} (0.6) & \textbf{48.6} (0.5) & \textbf{46.4} (0.2) & \textbf{48.6} (0.5) & \textbf{46.4} (0.2) \\
mPLUG-Owl2 & \textbf{47.5} (0.2) & \textbf{44.6} (0.3) & \textbf{47.8} (0.2) & \textbf{48.6} (0.7) & \textbf{49.8} (0.5) & \textbf{46.7} (0.2) & \textbf{49.8} (0.5) & \textbf{46.7} (0.2) \\
Idefics2 & \textbf{46.4} (0.2) & \textbf{41.7} (0.3) & \textbf{46.5} (0.2) & \textbf{45.2} (0.7) & \textbf{49.2} (0.5) & \textbf{44.8} (0.2) & \textbf{49.2} (0.5) & \textbf{44.8} (0.2) \\
Mantis-SigLIP & \textbf{46.7} (0.2) & \textbf{42.8} (0.3) & \textbf{46.7} (0.2) & \textbf{44.0} (0.6) & \textbf{48.9} (0.5) & \textbf{45.2} (0.2) & \textbf{48.9} (0.5) & \textbf{45.2} (0.2) \\
Mantis-Idefics2 & \textbf{46.4} (0.2) & \textbf{41.9} (0.3) & \textbf{46.3} (0.2) & \textbf{42.3} (0.6) & \textbf{48.4} (0.5) & \textbf{44.7} (0.2) & \textbf{48.4} (0.5) & \textbf{44.7} (0.2) \\
mPLUG-Owl3 & \textbf{46.3} (0.2) & \textbf{42.6} (0.3) & \textbf{46.3} (0.2) & \textbf{47.4} (0.7) & \textbf{49.7} (0.5) & \textbf{45.0} (0.2) & \textbf{49.7} (0.5) & \textbf{45.0} (0.2) \\

LLaVA-OV & \textbf{49.5} (0.2) &	\textbf{48.3} (0.3) &	\textbf{49.7} (0.2) &	\textbf{48.2} (0.7) &	\textbf{49.5} (0.5) &	\textbf{49.3} (0.2) &	\textbf{49.5} (0.5) &	\textbf{49.3} (0.2) \\
PaliGemma-2 & \textbf{35.4} (0.1) &	\textbf{40.6} (0.3) &	\textbf{32.5} (0.1) &	\textbf{33.0} (0.5) &	\textbf{33.1} (0.4) &	\textbf{35.6} (0.1) &	\textbf{33.1} (0.4) &	\textbf{35.6} (0.1) \\
PaliGemma-2-Mix & \textbf{47.5} (0.2) &	\textbf{42.0} (0.3) &	\textbf{48.0} (0.2) &	\textbf{44.1} (0.7) &	\textbf{47.1} (0.5) &	\textbf{46.1} (0.2) &	\textbf{47.1} (0.5) &	\textbf{46.1} (0.2) \\

        \bottomrule
    \end{tabular}
    \caption[]{SD and SEM scores across various models on our datasets in the zero-shot multiple choice scenario. SD scores are highlighted in bold, with SEM scores provided in parentheses. "SG" refers to scene graph. "VG" and "GLDv2" refer to Visual Genome and Google Landmarks Datasets v2 respectively.} 
    \label{table:std-multichoice}
\end{table*}

\end{document}